\documentclass{article}

\usepackage[preprint]{neurips_2026}
 \usepackage[preprint]{neurips_2026}

\usepackage[utf8]{inputenc} % allow utf-8 input
\usepackage[T1]{fontenc}    % use 8-bit T1 fonts
\usepackage{hyperref}       % hyperlinks
\usepackage{url}            % simple URL typesetting
\usepackage{booktabs}       % professional-quality tables
\usepackage{amsfonts}       % blackboard math symbols
\usepackage{nicefrac}       % compact symbols for 1/2, etc.
\usepackage{microtype}      % microtypography
\usepackage{xcolor}         % colors
\usepackage{graphicx}
\usepackage{float}
\usepackage{multirow}
\usepackage{amsmath}

\title{Hybrid Neural World Models}

\title{Hybrid Neural World Models}
\author{%
  Pranav Lakshmanan$^{*}$, Paras Chopra \\
  Lossfunk \\
  \texttt{pranav.lakshmanan@lossfunk.com}, \texttt{paras@lossfunk.com}
}
\date{May 2026}


\begin{document}

\centerline{\includegraphics[width=0.4\linewidth]{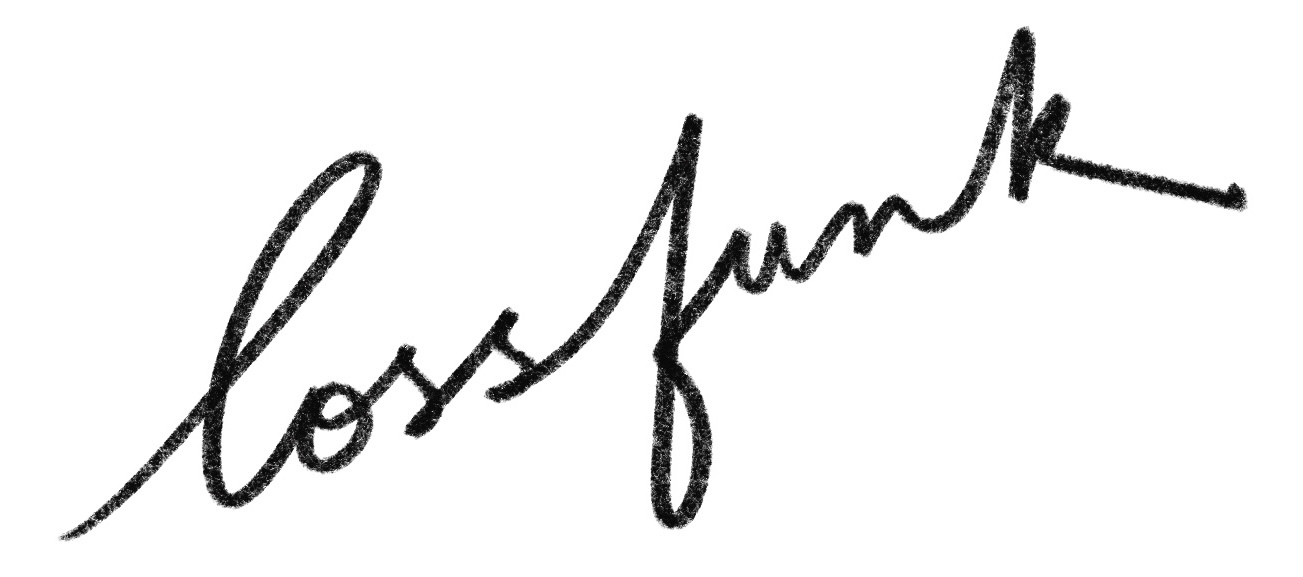}}
\maketitle

\begin{abstract}
  Neural surrogates promise large speedups over classical solvers for physical dynamics but fail silently at sharp dynamical events such as shocks, fronts, and contact. We present
  \textbf{hybrid neural world models} for physical dynamics: a recipe for training and deploying multi-horizon surrogates in physical state space, where a single network with
  continuous horizon conditioning is trained with direct supervision against textbook reference solvers to predict any future state at horizon $T$ in one forward pass. Although no part of the
   training data, loss function, or architecture supervises discontinuity location, the trained surrogate encodes it implicitly, recoverable from its forward passes alone as a per-trajectory
  \textbf{error map} that concentrates on shocks, fronts, and contacts, and stays small elsewhere. The map is competitive with or better than standard label-free baselines including deep
  ensembles, learned error heads, gradient-magnitude indicators, and locally-adaptive conformal prediction, while using only a single trained network and requiring no calibration set or
  governing-equation knowledge. The recipe supports two operating points. \textbf{Mode 1} runs the surrogate alone for maximum throughput, with same-hardware CPU speedups of $26\times$ to
  $72\times$ against textbook solvers on the PDE environments. \textbf{Mode 2} uses the error map to gate a reference-solver fallback, deferring uncertain trajectories and roughly halving the
   surrogate's residual error at the default operating point. The recipe applies without modification across reaction-diffusion, compressible Euler, and rigid-body collision dynamics.
\end{abstract}

  \section{Introduction}
  \label{sec:intro}

  Neural surrogates promise large speedups over reference solvers for physical dynamics, but they fail silently at sharp dynamical events such as shocks, fronts, and contact. A surrogate that
   returns a plausible field everywhere, even when the underlying physics is non-smooth, is unsafe to deploy in any pipeline that does not have access to a ground-truth simulator at
  inference. Detecting where the surrogate is unreliable, without solving the system from scratch, is the bottleneck for using these models at scale.

  We train a multi-horizon surrogate: a network with continuous horizon conditioning that predicts any future state at horizon $T$ in one forward pass. From this trained surrogate we
   extract two inference-time pieces at no additional cost. An error map compares the surrogate's prediction at horizon $T$ against its iterated prediction at half-horizon; disagreement
  between the two is large where the surrogate is unreliable. A two-mode deployment policy then uses Mode~1 (surrogate alone) for routine trajectories and Mode~2 (selective solver fallback
  gated by the error map) for trajectories the map flags.

  We validate this across three physical systems spanning two regimes: Oregonator (a reaction-diffusion PDE with propagating chemical fronts), Euler 2D (compressible flow PDE with shock
  formation), and Ball 3D (a rigid-body ODE with collision events). The three together test whether the same recipe holds across continuous-field PDE physics and discrete-event ODE dynamics
  without modification.

\begin{figure}[t]
  \centering
  \includegraphics[width=\linewidth]{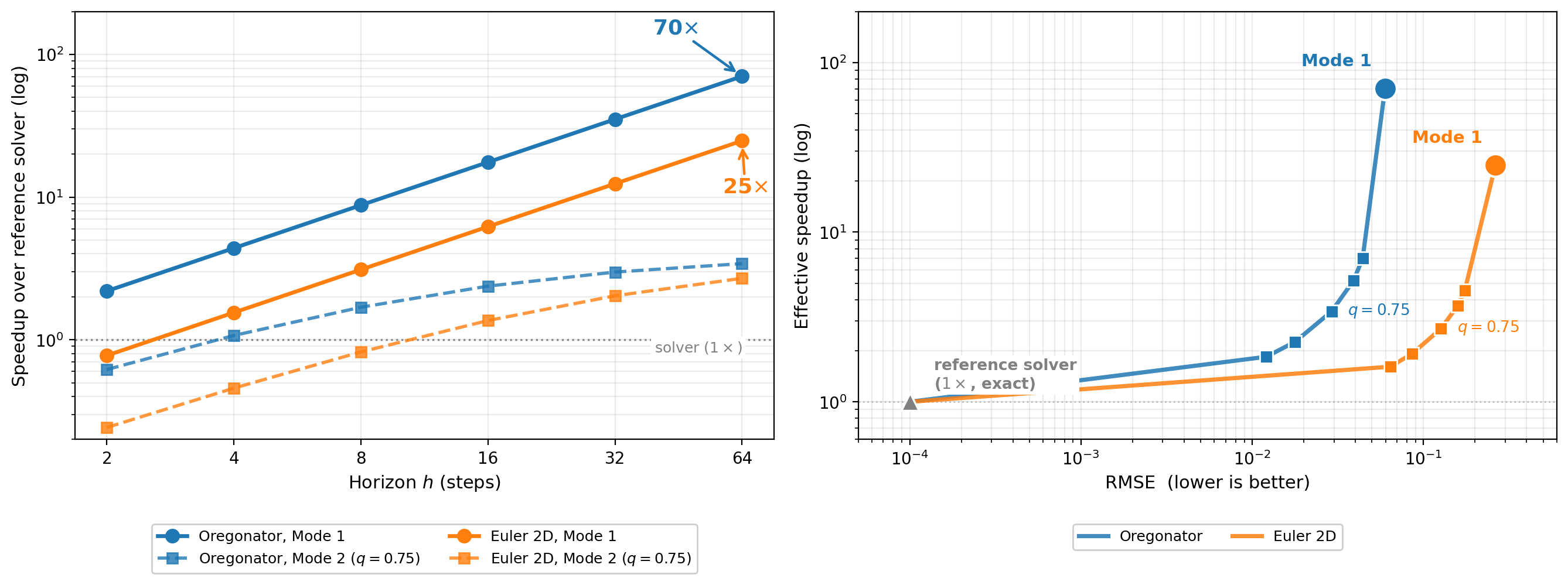}
  \caption{\textbf{Wall-clock speed comparison for hybrid neural world models.} (a) CPU-CPU wall-clock speedup vs horizon for Mode 1 (surrogate alone, solid) and Mode 2 (with trust-aware
  fallback at $q{=}0.75$, dashed); the two PDE environments reach $25\times$ and $70\times$ at $h{=}64$. (b) Pareto frontier in (RMSE, speedup) space at $h{=}64$. Each curve runs from Mode 1
  (top circle, fastest) through Mode 2 $q$ values (squares; $q \in \{0.9, 0.85, 0.75, 0.6, 0.5\}$) to the reference solver (bottom-left triangle, exact). Here $q$ is the surrogate-keep
  fraction: the error map defers the remaining $(1-q)$ of trajectories to the reference solver.}
  \label{fig:speedup}
  \end{figure}

  Concurrent hybrid neural-classical solvers \citep{roy2025anchor,srikishan2025hyper} use gate solver invocation via PDE residuals or reinforcement-learning policies; horizon-conditioned
  multi-step surrogates \citep{nguyen2024stormer,bi2023pangu,herde2024poseidon} use multi-step predictions for accuracy rather than as trust signals. Our error map needs neither and applies
  in identical form to PDE and ODE physics.

  \vspace{5pt}

\paragraph{Contributions.}
  \begin{enumerate}
\item A training recipe for multi-horizon shortcut surrogates in physical state space: continuous horizon conditioning via FiLM, direct supervised regression against reference-solver
  outputs, and a $10\%$ DAgger refinement. Appendix~\ref{app:self_consistency} shows the self-consistency loss of \citet{frans2024shortcut}, when ported from diffusion to physical state,
  collapses to the identity map; direct supervision is therefore necessary, not just preferable.
  \item An inference-time error map computed from the trained surrogate alone, with no extra training, no calibration set, and no governing-equation knowledge, that ranks trajectories by
  their true error and outperforms label-free baselines including deep ensembles, learned error heads, gradient-magnitude indicators, and locally-adaptive conformal prediction.
  \item A two-mode deployment policy validated across reaction-diffusion, compressible Euler, and rigid-body collision dynamics. Mode~1 runs the surrogate alone, producing same-hardware CPU
  speedups of $26\times$ to $72\times$ against textbook solvers on the PDE environments. Mode~2 uses the error map to gate a reference-solver fallback, roughly halving the surrogate's
  residual error at the default operating point.
  \end{enumerate}

  \section{Related Work}
  \label{sec:related}

  \paragraph{Multi-horizon and multi-scale neural surrogates.}
  A growing body of work trains a single neural surrogate at multiple lead times for prediction accuracy: Stormer \citep{nguyen2024stormer}, Pangu-Weather \citep{bi2023pangu}, and Poseidon
  \citep{herde2024poseidon} train atmospheric transformers across forecast horizons; ShockCast \citep{helwig2025shockcast} predicts adaptive timestep sizes for high-speed flows; hierarchical
  multiscale time-steppers \citep{liu2022hits,hamid2024ahits} stack and switch networks at fixed timescales; TI-DeepONet \citep{nayak2025tideeponet} learns integration coefficients for stable
   rollouts. None of these extract a per-trajectory uncertainty estimate. We use the same training pattern to read out the disagreement between predictions at horizons $T$ and $T/2$ as a
  label-free error map.

  \paragraph{Amortised multi-step prediction.}
  A separate thread amortises multiple sub-steps into a single forward pass. Shortcut models \citep{frans2024shortcut} apply this to diffusion sampling with an optional self-consistency loss;
   Dreamer 4 \citep{hafner2025dreamer4} extends similar ideas to action-conditioned latent video world models, in the lineage of \citet{ha2018world}. We work in physical state space with
  ground-truth solver supervision at training time, so we supervise directly rather than through self-consistency, which is actively harmful here: Appendix~\ref{app:self_consistency} shows it
   collapses to the identity map. Existing models use multi-horizon prediction only as a forward-pass speedup; we additionally extract a label-free trust signal
  (Section~\ref{sec:method:errormap}) that gates selective solver fallback.

  \paragraph{Trust signals and hybrid neural-classical schemes.}
  Hybrid neural-classical solvers gate solver invocation by an uncertainty signal: ANCHOR \citep{roy2025anchor} uses a PDE-residual EMA, HyPER \citep{srikishan2025hyper} trains an RL policy,
  PDE-Refiner \citep{lippe2023pderefiner} requires stochastic sampling at inference, Calibrated PI-UQ \citep{gopakumar2025calibrated} couples residuals with a held-out conformal set,
  cycle-consistency UQ \citep{huang2024cycle} closes a forward-backward loop, and \citet{beck2020shock} train a supervised CNN on labelled shocks. Each requires what we do not: a residual, a
  learned policy, a calibration set, or labels. Our probe falls out of the multi-horizon training already done, and applies to PDE and rigid-body environments without modification.

  \paragraph{Classical UQ baselines and selective prediction.}
  Deep ensembles \citep{lakshminarayanan2017simple,ovadia2019can} remain the standard label-free UQ baseline but cost $K\times$ to train and store; we match a $K{=}3$ ensemble at one-third
  the cost. Conformal prediction \citep{romano2019conformalized,bostroem2017accelerating,tibshirani2019conformal,angelopoulos2023conformal} gives distribution-free intervals but needs a
  calibration set and degrades under shift. Selective prediction \citep{geifman2017selective} and learning-to-defer \citep{mozannar2020consistent,verma2023learning} formalise trust-aware
  decisions; Mode~2 fits this framework with the error map as deferral score and no separately trained defer rule.

  \paragraph{Numerical analysis lineage.}
  Embedded Runge-Kutta methods \citep{dormand1980family} and Richardson extrapolation \citep{richardson1911} estimate local truncation error by comparing two solver legs of \emph{known order
  of convergence}. Our probe borrows the two-leg structure but assumes only that multi-horizon supervised training has imparted internal coherence on smooth dynamics, not a known order.

  \paragraph{Foundational neural surrogates.}
  Fourier neural operators \citep{li2021fno}, DeepONet \citep{lu2021deeponet}, and physics-informed networks \citep{raissi2019pinns} are the architectural primitives that motivated
  neural-replacement solvers; PDEBench \citep{takamoto2022pdebench} benchmarks them. Our contribution is orthogonal: a deployment-time trust signal for any sufficiently fast surrogate, not a
  new architecture.

  \section{Method}
  \label{sec:method}

  \subsection{Multi-horizon shortcut surrogate}
  \label{sec:method:surrogate}

  Let $s_t \in \mathcal{S}$ be the physical state at time $t$ and $\Phi_T : \mathcal{S} \to \mathcal{S}$ the ground-truth flow map that advances the state by horizon $T$. We train a single
  neural network $f_\theta(s, T) \approx \Phi_T(s)$ that predicts the future state at any $T$ in one forward pass.

  \paragraph{Architecture.}
  The recipe is architecture-agnostic. The error map and the two-mode policy work with any network that takes a state and a horizon as input and produces a state of the same shape, regardless
   of how the conditioning is implemented. In practice we use the conventional choice for each domain: a U-Net \citep{ronneberger2015unet} for two-dimensional grid-structured PDE fields,
  where its multiscale receptive field matches the spatial structure of fronts and shocks, and a residual MLP for low-dimensional state vectors. The horizon $T$ is encoded as a continuous
  embedding via FiLM \citep{perez2018film}; we use FiLM rather than AdaLN, concatenation, or cross-attention because it is parameter-light and slots cleanly into both architectures without
  redesign. None of the analysis depends on this choice.

  \paragraph{Training.}
  Training pairs $(s_0, s_T, T)$ are drawn from full-resolution simulator rollouts, with $T$ sampled uniformly from the set $\{2, 4, 8, 16, 32, 64\}$. We chose this geometric progression for
  two reasons. It covers two orders of magnitude in horizon with only six values, which keeps the loss well balanced across scales without needing per-horizon weighting. And the doubling
  structure ensures every horizon $T$ in the set has its half-horizon $T/2$ also in the set, which the inference probe in Section~\ref{sec:method:errormap} requires. The training objective is
   direct supervised regression in normalised state space:
  \begin{equation}
  \mathcal{L}_{\text{sup}}(\theta) = \mathbb{E}_{(s_0, s_T, T)}\,\bigl\|\tilde{f}_\theta(s_0, T) - \tilde{s}_T\bigr\|_2^2,
  \label{eq:supervised}
  \end{equation}
  where $\tilde{\cdot}$ denotes per-channel normalisation by training statistics. We supervise directly rather than via a self-consistency loss because the latter collapses to the identity
  map in physical state space (Appendix~\ref{app:self_consistency}).

 \subsection{Error map at inference}
  \label{sec:method:errormap}

  At inference time, we ask the network the same prediction question two ways: once in a single forward pass at horizon $T$, and once by chaining two forward passes at half-horizon $T/2$. The
   error map is the magnitude of the disagreement between these two answers:
  \begin{equation}
  \hat{e}(s, T) \;=\; \bigl\|\,\underbrace{f_\theta(s, T)}_{\text{single-shot at } T} \;-\; \underbrace{f_\theta\!\bigl(f_\theta(s, T/2),\, T/2\bigr)}_{\text{chained at } T/2}\,\bigr\|_2.
  \label{eq:errormap}
  \end{equation}
  For spatial state, $\hat{e}$ is computed per cell (the norm is taken across channels), producing a heatmap over the domain that is large where the surrogate is unreliable and small
  elsewhere. For low-dimensional state, $\hat{e}$ is a scalar per trajectory.

  \paragraph{Why this works.}
  The true flow $\Phi$ satisfies the semigroup composition property $\Phi_{T+S} = \Phi_T \circ \Phi_S$ \citep{engel2000oneparam}; in particular $\Phi_T = \Phi_{T/2} \circ \Phi_{T/2}$, and
  Eq.~\ref{eq:errormap} measures how closely the trained surrogate inherits this identity. Multi-horizon supervised training (Eq.~\ref{eq:supervised}) drives $f_\theta$ toward $\Phi_{T'}$ at
  every $T'$ in the ladder. On a region $K \subset \mathcal{S}$ where the per-horizon error is uniformly bounded by $\varepsilon$, $\Phi_{T/2}$ is $L$-Lipschitz, and chained predictions stay
  in $K$, two applications of the triangle inequality yield $\hat{e}(s, T) \le \varepsilon\,(2 + L)$ (Appendix~\ref{app:bound}). The bound is directional only: it forces $\hat{e}$ to be small
   on smooth regions but is vacuous at shocks and contacts where $\Phi_{T/2}$ is discontinuous;  empirical confirmation that $\hat{e}$ correspondingly grows at non-smooth features is provided by the per-cell error maps in Figures~\ref{fig:emap_oreg}--\ref{fig:emap_ball3d} and the AUROC
   values in Table~\ref{tab:auroc_sd_coverage}. Training at a single horizon does not deliver the per-horizon error bound at $T/2$ and the probe degenerates to noise
  (Appendix~\ref{app:single_horizon}).

  \paragraph{Distinctions and caveats.}
  Classical Richardson extrapolation \citep{dormand1980family,richardson1911} compares two solver legs of \emph{known order of convergence}; we require no such assumption
  (Appendix~\ref{app:richardson}). Shortcut models \citep{frans2024shortcut} train against a self-consistency loss with an exact identity-map solution in physical state space and collapse to
  it (Appendix~\ref{app:self_consistency}); we instead read the same residual after supervised training has driven $f_\theta$ toward $\Phi$. The argument is informal: the partition of
  $\mathcal{S}$ into smooth regions is not given a priori, the bound is upper-only, and $L$ is environment-specific.
  
  \subsection{Two modes of inference}
  \label{sec:method:modes}

  \paragraph{Mode 1 (surrogate alone).}
  $\hat{s}_T = f_\theta(s_0, T)$ in a single forward pass. Maximum throughput; the surrogate's errors at sharp events are accepted as the cost.

  \paragraph{Mode 2 (trust-aware fallback).}
  For each trajectory we aggregate $\hat{e}$ to a scalar $\bar{e}$ (spatial mean for spatial state, $\hat{e}$ itself for low-dimensional state). Trajectories with $\bar{e} \le \tau$ keep the
  surrogate prediction; the rest are handed to the reference solver.

  \paragraph{Setting the threshold.}
  The threshold $\tau$ is parametrised by a single deployment knob $q \in [0, 1]$, the surrogate-keep fraction; $\tau$ is the $q$-quantile of $\bar{e}$ on a held-out validation set, never on
  test data. Smaller $q$ defers more trajectories (slower, more accurate); larger $q$ keeps more on the surrogate. We use $q{=}0.75$ as the default; the full sweep is in
  Section~\ref{sec:experiments}.

  \paragraph{Trajectory-level deferral.}
  $\hat{e}$ is useful at two scales: a trajectory-mean $\bar{\hat{e}}$ that gates Mode~2, and a per-cell field that visualises where the surrogate is unreliable (Section~\ref{sec:5_trust},
  Figures~\ref{fig:emap_oreg}--\ref{fig:emap_ball3d}). We deploy with the trajectory-mean because reference solvers integrate stencils across cell boundaries (a partial-domain handoff would
  need custom boundary coupling), and because the surrogate's forward pass at horizon $T$ entangles all cells through its receptive field. The per-cell field is preserved at deployment and
  remains available for adaptive mesh refinement or domain-aware tooling.

  \section{Environments}
  \label{sec:envs}

  We test the recipe on three physical systems: two PDEs of broad practical interest, and one ODE that tests whether it transfers beyond PDE physics.

  \subsection{Oregonator: reaction-diffusion PDE}
  The Oregonator \citep{field1972oregonator,tyson1980oregonator} is a two-variable reaction-diffusion model of the Belousov--Zhabotinsky oscillating chemical reaction; the same mathematical
  structure governs cardiac action potentials and neural excitation waves. The state is a two-channel concentration field $u(x, y, t) \in \mathbb{R}^{2 \times 256 \times 256}$ on a periodic
  grid, governed by
  \[
  \partial_t u = D \nabla^2 u + R(u),
  \]
  with $D$ the diffusion rate and $R(\cdot)$ the Tyson reduction of the reaction kinetics. Dynamics are smooth between fronts and sharp at the front itself.

  \subsection{Euler 2D: compressible flow PDE}
  The 2D compressible Euler equations govern inviscid mass--momentum--energy conservation in supersonic aerodynamics, blast waves, and astrophysical jets. The state is a four-channel
  conserved-variable field $\mathbf{q} = (\rho, \rho u, \rho v, E)$ on a $128 \times 128$ grid, satisfying
  \[
  \partial_t \mathbf{q} + \nabla \cdot \mathbf{F}(\mathbf{q}) = 0,
  \]
  with the standard ideal-gas closure relating energy to pressure. The dynamics produce shocks, contact discontinuities, and smooth rarefactions, often interacting.

  \subsection{Ball 3D: rigid-body collision ODE}
  A ball bounces inside a unit cube under gravity with elastic-with-loss wall collisions; practical analogues include robotics contact and granular flow. The state is a $9$-vector of
  position, linear velocity, and angular velocity. Between collisions $\ddot{\mathbf{x}} = \mathbf{g}$ and $\dot{\boldsymbol{\omega}} = 0$; at a wall the normal velocity component is
  reflected with restitution $\epsilon$. Because the state is non-spatial, the error map collapses to a scalar per trajectory.

  \section{Experiments and Results}
  \label{sec:experiments}

  We report five experiments establishing the paper's claims:
  (i) the surrogate trains to usable accuracy on each environment;
  (ii) step-doubling is a label-free trust signal that ranks trajectories by their true error;
  (iii) it outperforms label-free baselines (deep ensembles, learned error heads, random TTA, gradient magnitude, locally adaptive conformal prediction);
  (iv) Mode~1 produces a meaningful speedup over the reference solver;
  (v) Mode~2 cuts the surrogate's residual error at a controlled speedup cost.
  Closed-loop rollout, single-horizon and DAgger weight ablations, the self-consistency-only training collapse, beyond-$T_{\max}$ extrapolation, the per-quantile $q$-sweep, cross-seed AUROC
  stability, energy and momentum baselines for Ball~3D, and per-horizon Mode~1 vs Mode~2 RMSE across all distribution splits are reported in Appendix~\ref{app:additional_experiments}.

  \begin{table}[H]
  \centering
  \caption{Dataset and architecture per environment. Training horizon ladder $h \in \{1, 2, 4, 8, 16, 32, 64\}$ for all environments. Backbones are dictated by data shape, not by tuning.}
  \label{tab:dataset_spec}
  \small
  \setlength{\tabcolsep}{6pt}
  \renewcommand{\arraystretch}{1.05}
  \begin{tabular}{l c c c}
  \toprule
   & Oregonator & Euler 2D & Ball 3D \\
  \midrule
  State shape           & $256{\times}256{\times}2$ & $128{\times}128{\times}4$ & $\mathbb{R}^9$ \\
  Trajectory length $T$ & $201$  & $100$  & $101$ \\
  Base step $\Delta t$  & $0.05$ & $2\mathrm{e}{-3}$ & $0.01$ \\
  Train trajectories    & $1200$ & $500$  & $1000$ \\
  Val / test            & $150 / 150$ & $100 / 100$ & $200 / 200$ \\
  OOD-near / OOD-far    & $250 / 250$ & $150 / 150$ & $200 / 200$ \\
  \midrule
  Backbone              & U-Net & U-Net & MLP + FiLM \\
  Parameters            & $3.56$\,M & $3.56$\,M & $0.69$\,M \\
  Final val MSE         & $0.104$ & $0.016$ & $0.024$ \\
  \bottomrule
  \end{tabular}
  \end{table}

  \subsection{Per-environment training and reference solvers}
  \label{sec:5_setup}

  \paragraph{Reference solvers.}
  \textit{Oregonator}: Strang-split integrator on a periodic grid; implicit Euler for the stiff reaction term, explicit FTCS for diffusion with internal CFL substepping; Tyson parameters
  $(\varepsilon, q, f, D) = (0.05, 0.002, 2.0, 1.0)$. \textit{Euler 2D}: finite-volume HLL Riemann solver \citep{harten1983hll} with $\mathrm{CFL}{=}0.4$ and $\gamma{=}1.4$. \textit{Ball 3D}:
   pure-NumPy rigid-body integrator with semi-implicit Euler at $50$ substeps per saved frame and analytic axis-aligned wall collisions; per-trajectory we sample gravity $g \in [-10.5,
  -9.0]\,\mathrm{m/s}^2$, restitution $e \in [0.7, 0.95]$, $\|v_0\| \in [1, 3]\,\mathrm{m/s}$, $\|\boldsymbol{\omega}_0\|_\infty \le 5\,\mathrm{rad/s}$.

  \paragraph{Surrogates and training.}

  PDE surrogates use the standard U-Net of Section~\ref{sec:method}; the Ball~3D surrogate is a four-block FiLM-conditioned residual MLP.
  Parameter counts are in Table~\ref{tab:dataset_spec}; backbone is dictated by data shape and we make no environment-specific architectural changes.All three surrogates are trained with the same recipe: AdamW, multi-horizon supervision over $h \in \{1, 2, 4, 8, 16, 32, 64\}$, and a $10\%$ DAgger \citep{ross2011dagger} refinement
  against the reference solver. The
  trust signal and speedup behaviour reported below are properties of this training recipe, not of any chosen network.

  \subsection{The trust signal ranks trajectories by their true error}
  \label{sec:5_trust}

 Table~\ref{tab:auroc_sd_coverage} reports step-doubling AUROC against true error across all $54$ cells (three environments, six training horizons, three distribution splits). The signal
  gives useful ranking (AUROC ${>}0.5$) in $53$ of $54$ cells, with median AUROC $0.76$ and $50$ cells above $0.65$. Two regimes deserve closer reading. The Oregonator AUROC saturates at
  $0.65$--$0.77$ because the surrogate is nearly uniformly accurate on this environment (M1 RMSE $0.06$ at $h{=}64$), leaving little spread to discriminate; this is a structural ceiling
  rather than a failure mode, and Mode~2 still recovers $40$--$68\%$ of the residual RMSE on these cells (Appendix~\ref{app:m1m2_horizons}). The Euler AUROC rises toward $1.0$ under
  distribution shift because surrogate failures become bimodally distributed and easy to flag. The one cell where the probe genuinely fails is Ball~3D under far-OOD restitution and gravity at
   $h{=}16$, where AUROC drops to $0.44$ (below chance): at extreme restitution the position error is driven by accumulated wall-collision miscount rather than per-step extrapolation, and the
   surrogate's failure mode no longer tracks step-size sensitivity. Cross-seed analysis (Appendix~\ref{app:cross_seed}) confirms this is not a single-seed artefact ($0.43 \pm 0.02$ across
  three seeds); Section~\ref{sec:limitations} discusses the regime where the recipe breaks down.

  \begin{table}[H]
  \centering
  \caption{Per-cell AUROC at the $75$th-percentile error threshold, $80{-}100$ pairs per cell.}
  \label{tab:auroc_sd_coverage}
  \small
  \setlength{\tabcolsep}{4pt}
  \begin{tabular}{c | ccc | ccc | ccc}
  \toprule
   & \multicolumn{3}{c|}{Oregonator} & \multicolumn{3}{c|}{Euler 2D} & \multicolumn{3}{c}{Ball 3D} \\
  $h$ & test & OOD-n & OOD-f & test & OOD-n & OOD-f & test & OOD-n & OOD-f \\
  \midrule
   2 & 0.70 & 0.68 & 0.65 & 0.97 & 0.90 & 1.00 & 0.90 & 0.95 & 0.88 \\
   4 & 0.70 & 0.71 & 0.69 & 0.97 & 0.94 & 1.00 & 0.78 & 0.96 & 0.68 \\
   8 & 0.72 & 0.72 & 0.72 & 0.95 & 0.93 & 0.99 & 0.76 & 0.83 & 0.76 \\
  16 & 0.77 & 0.74 & 0.72 & 0.92 & 0.82 & 0.95 & 0.76 & 0.82 & 0.44 \\
  32 & 0.72 & 0.73 & 0.67 & 0.98 & 0.92 & 0.99 & 0.75 & 0.68 & 0.59 \\
  64 & 0.75 & 0.65 & 0.64 & 0.95 & 0.98 & 0.98 & 0.76 & 0.71 & 0.55 \\
  \bottomrule
  \end{tabular}
  \end{table}

\begin{figure}[H]
    \centering
    \includegraphics[width=\linewidth]{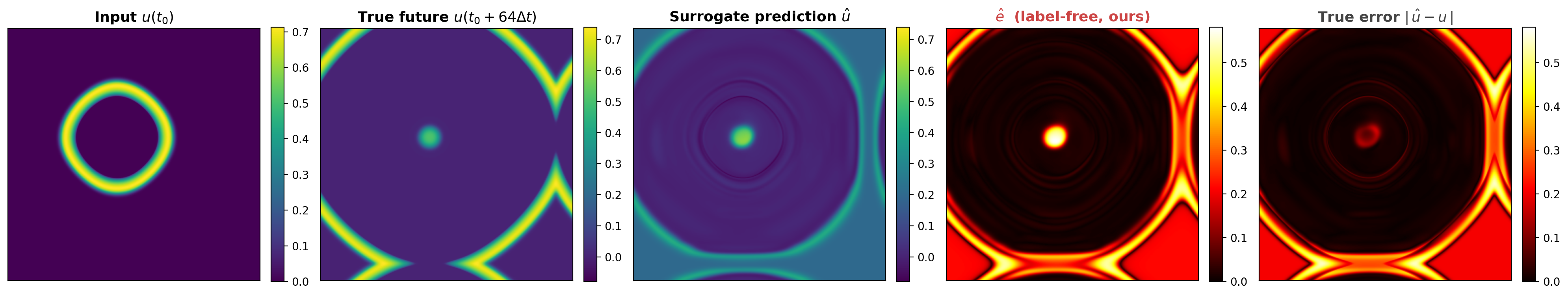}
    \caption{\textbf{Oregonator visual proof.} Input $u(t_0)$, true future at $t_0 + 64\Delta t$, surrogate prediction $\hat{u}$, label-free error map $\hat{e}$, and true per-cell error
  $|\hat{u} - u|$. Two right panels share a colour scale.}
    \label{fig:emap_oreg}
  \end{figure}

\begin{figure}[H]
    \centering
    \includegraphics[width=\linewidth]{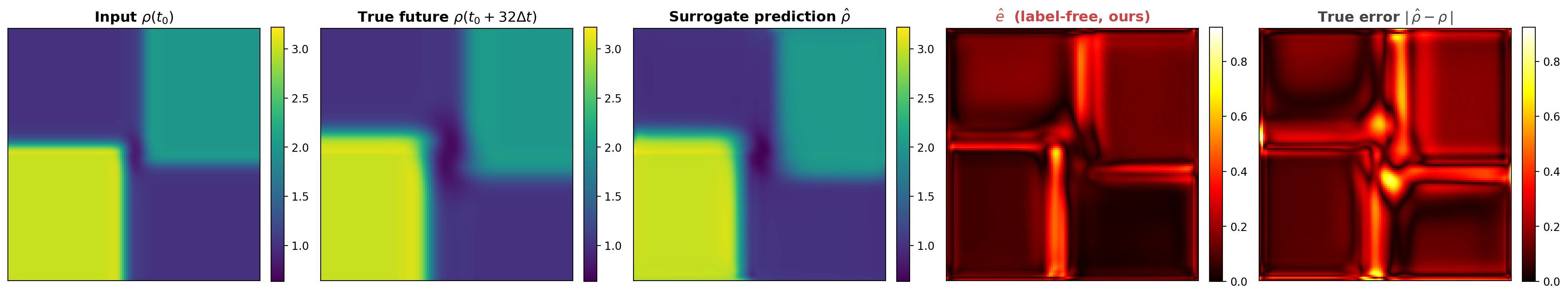}
    \caption{\textbf{Euler 2D visual proof, Schulz--Rinne quadrant configuration.} Same five-panel layout as Figure~\ref{fig:emap_oreg}, density channel $\rho$.}
    \label{fig:emap_euler}
  \end{figure}

\begin{figure}[H]
    \centering
    \includegraphics[width=\linewidth]{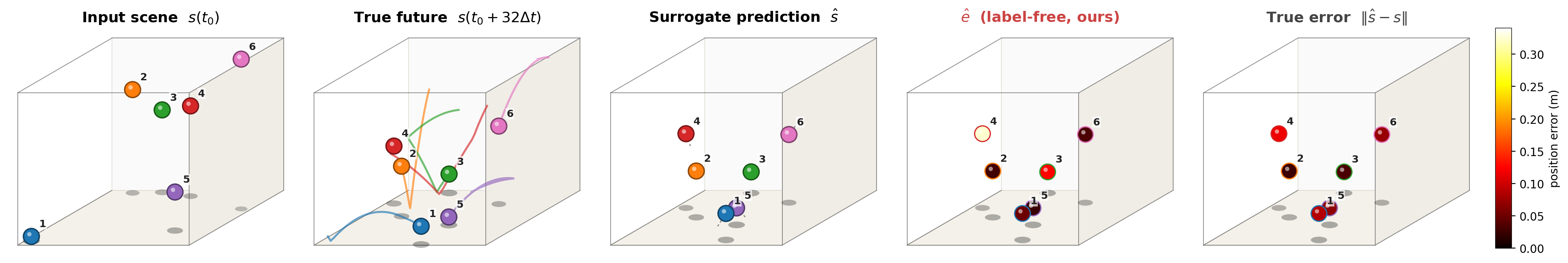}
    \caption{\textbf{Ball 3D visual proof, multi-ball composite scene.} Six independent ball trajectories in a shared isometric view. Cols 1--3: input, true future, and surrogate prediction
  at $t_0 + 32\Delta t$ with identity colours. Cols 4--5: same predicted positions recoloured by per-ball $\hat{e}$ and true error.}
    \label{fig:emap_ball3d}
  \end{figure}

  \paragraph{Spatial alignment and physics-aware selectivity.}
  Figures~\ref{fig:emap_oreg},~\ref{fig:emap_euler}, and~\ref{fig:emap_ball3d} compare $\hat{e}$ against the true per-cell error $|\hat{s}-s|$ on representative trajectories. The two fields
  agree closely: $\hat{e}$ concentrates on the propagating reaction front in Oregonator, on the four contact discontinuities of the Schulz--Rinne configuration in Euler, and on the
  trajectories that develop the largest position error in Ball~3D, with no per-cell supervision in training. Three pieces of evidence rule out the account that $\hat{e}$ is a generic edge
  detector or a proxy for prediction-error magnitude. First, $\hat{e}$ stays dark on the smooth quadrant interiors of the Schulz--Rinne configuration that any image-domain edge detector would
   treat as identical to the contact discontinuities. Second, the same selectivity holds across U-Net (PDE) and MLP (Ball~3D) backbones, consistent with the signal being a property of the
  training recipe rather than architectural inductive bias. Third, the single-horizon ablation in Appendix~\ref{app:single_horizon} shows that a surrogate with comparable prediction error has
   a structurally undefined probe, and classical Richardson extrapolation (Appendix~\ref{app:richardson}), which directly estimates prediction-difficulty under truncation-order assumptions,
  fails at $h \ge 4$ where step-doubling holds at AUROC $0.81$--$0.97$.

  \subsection{Step-doubling outperforms label-free baselines}
  \label{sec:5_baselines}

  Table~\ref{tab:auroc_baselines} compares step-doubling against five label-free baselines on the test split at $h \in \{16, 32, 64\}$. Step-doubling has the highest mean AUROC across the
  nine cells ($0.82$), and is the only signal applicable to all three environments without per-environment training or calibration. The closest competitor is the deep ensemble at mean $0.77$,
   which requires training $K{=}3$ independent networks rather than reusing one forward pass. Per-environment baselines (learned error head, conformal prediction) are competitive on the
  environment they were calibrated for but degrade or become inapplicable elsewhere; gradient magnitude is undefined on Ball~3D's $9$-dimensional state. Random TTA collapses on Euler
  ($\text{AUROC} < 0.4$), where small input perturbations produce predictions whose disagreement is anti-correlated with true error.

  \begin{table}[H]
  \centering
  \caption{Head-to-head AUROC against true error on test split. ``---'' marks combinations not applicable on the indicated environment. Bold marks the highest value per column.}
  \label{tab:auroc_baselines}
  \small
  \setlength{\tabcolsep}{4pt}
  \begin{tabular}{l | ccc | ccc | ccc}
  \toprule
  Method & \multicolumn{3}{c|}{Oregonator} & \multicolumn{3}{c|}{Euler 2D} & \multicolumn{3}{c}{Ball 3D} \\
   & $h{=}16$ & $h{=}32$ & $h{=}64$ & $h{=}16$ & $h{=}32$ & $h{=}64$ & $h{=}16$ & $h{=}32$ & $h{=}64$ \\
  \midrule
  Random TTA & 0.60 & 0.63 & 0.61 & 0.12 & 0.09 & 0.38 & 0.66 & 0.72 & 0.63 \\
  Gradient magnitude & 0.53 & 0.32 & 0.49 & 0.93 & 0.85 & \textbf{0.96} & --- & --- & --- \\
  Learned error head & \textbf{0.79} & 0.72 & 0.73 & 0.86 & 0.89 & 0.85 & 0.65 & 0.77 & \textbf{0.77} \\
  Locally adaptive CP & 0.56 & 0.72 & 0.63 & \textbf{0.93} & 0.89 & 0.91 & 0.47 & 0.59 & 0.47 \\
  Deep ensemble, $K{=}3$ & 0.75 & \textbf{0.73} & 0.71 & 0.81 & 0.82 & 0.83 & 0.74 & \textbf{0.81} & 0.70 \\
  \textbf{Step-doubling (ours)} & 0.77 & 0.72 & \textbf{0.75} & 0.92 & \textbf{0.98} & 0.95 & \textbf{0.76} & 0.75 & 0.76 \\
  \bottomrule
  \end{tabular}
  \end{table}

\subsection{Mode 1: surrogate-only deployment speedup}
  \label{sec:5_mode1}

\begin{figure}[H]
    \centering
    \includegraphics[width=\linewidth]{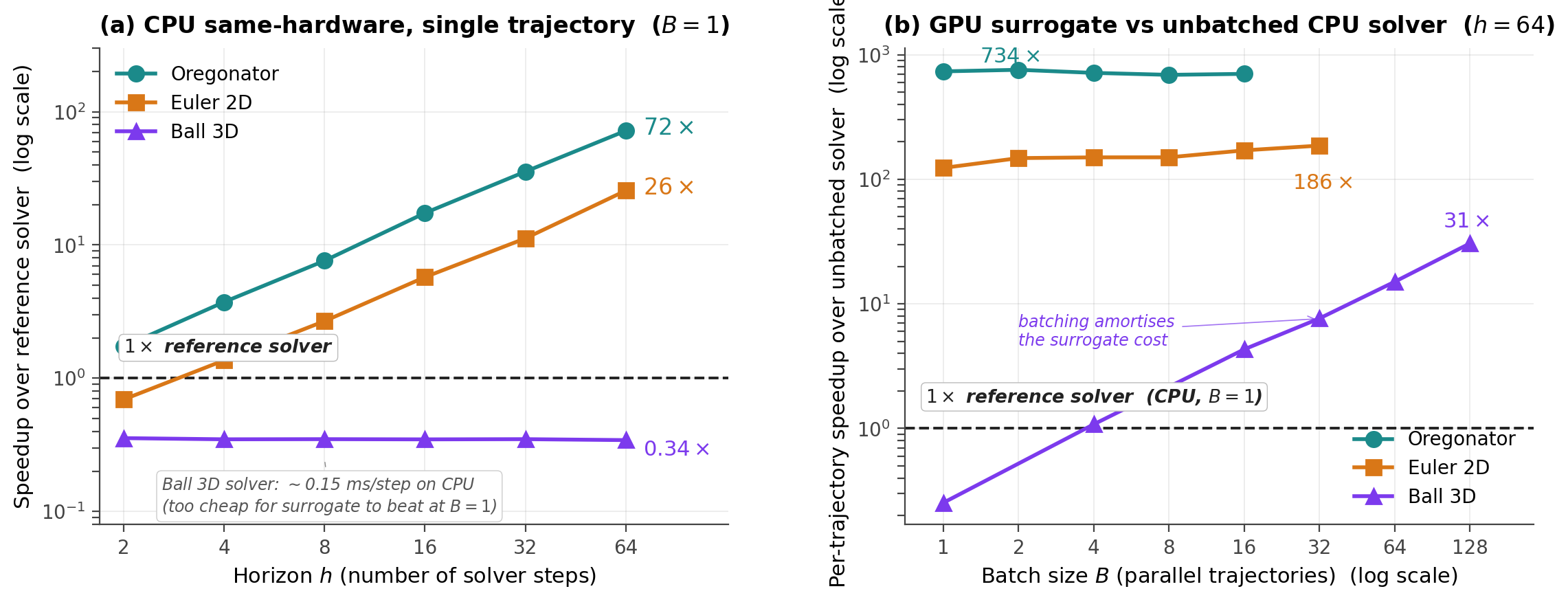}
    \caption{\textbf{Mode 1 deployment speedup.} (a) CPU same-hardware speedup over reference solver vs horizon at $B{=}1$. (b) GPU surrogate vs unbatched CPU solver at $h{=}64$ across batch
  sizes. Dashed line marks the reference solver baseline.}
    \label{fig:mode1}
  \end{figure}

  Figure~\ref{fig:mode1} reports Mode~1 wall-clock speedup against the reference solver. On a single trajectory at $B{=}1$ on the same CPU as the solver (panel~a), the surrogate's cost is
  invariant in horizon while the solver scales linearly, giving $72\times$ speedup on Oregonator and $26\times$ on Euler 2D at $h{=}64$. Ball~3D's pure-NumPy collision integrator is
  sub-millisecond per step, leaving no room for a single-call surrogate to win at $B{=}1$. On GPU at $h{=}64$ across batch sizes (panel~b), the PDE surrogates saturate near $B{=}1$ at
  $734\times$ on Oregonator and $186\times$ on Euler against the unbatched CPU reference; Ball~3D recovers from below-baseline at $B{=}1$ to $31\times$ at $B{=}128$, which is the regime in
  which many ball trajectories run in parallel.  Practitioners running these at scale would typically use JAX/JIT or GPU-vectorised solvers; panel~(a) is a clean compute-per-step comparison, not an upper bound on deployment speedup, and
  the trust-signal results below are independent of either choice.

  \subsection{Mode 2: gated deferral cuts surrogate RMSE}
  \label{sec:5_mode2}

\begin{figure}[H]
    \centering
    \includegraphics[width=0.62\linewidth]{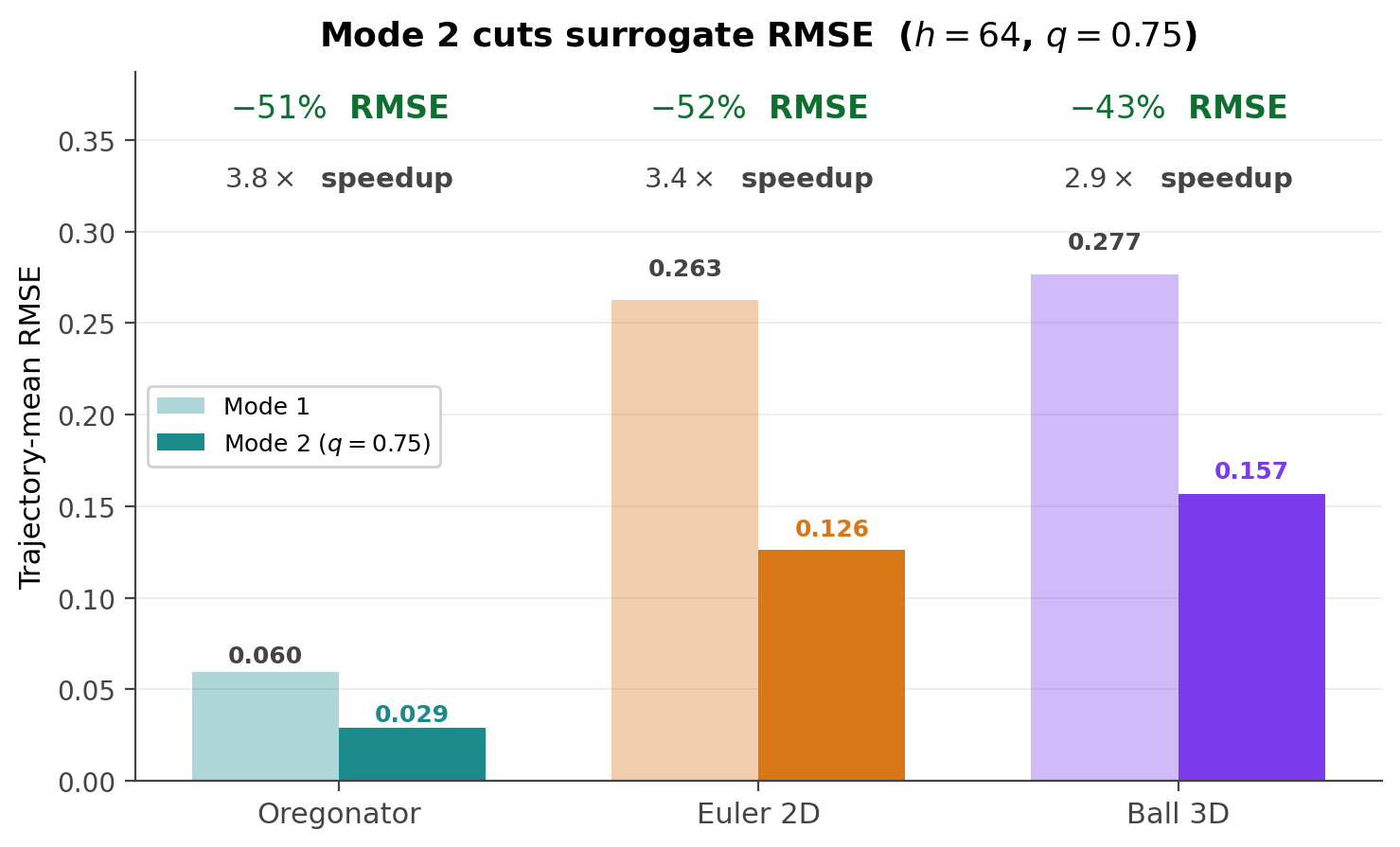}
    \caption{\textbf{Mode 2 cuts surrogate RMSE} at $h{=}64$, $q{=}0.75$. Mode 1 (faded) and Mode 2 (solid) trajectory-mean RMSE per environment; green annotations show relative reduction and
   the effective speedup retained.}
    \label{fig:mode2}
  \end{figure}

  Figure~\ref{fig:mode2} reports Mode~2 RMSE at the default $q{=}0.75$ on the test split at $h{=}64$. Across all three environments, gating the top $25\%$ riskiest trajectories to the
  reference solver cuts trajectory-mean RMSE by $43\%{-}52\%$ relative to the surrogate-alone baseline, while retaining a $\sim 3\times$ effective speedup. The reduction is monotone in $q$:
  smaller $q$ defers more trajectories and recovers more accuracy at higher cost; larger $q$ does the opposite. Per-horizon and per-distribution-shift breakdowns of the same comparison across
   $54$ cells (3 environments $\times$ 6 horizons $\times$ 3 splits) are reported in Appendix~\ref{app:m1m2_horizons}; every cell shows reduction, with the largest cuts ($-89\%$ on Euler
  far-OOD at $h{=}8$) occurring exactly where the surrogate fails worst.

  \section{Limitations}
  \label{sec:limitations}

  \paragraph{Solver-vectorisation caveat.}
  The high speedups in panel~(b) of Figure~\ref{fig:mode1} ($734\times$ on Oregonator, $186\times$ on Euler at $h{=}64$) compare a GPU-batched neural surrogate against an \emph{unbatched}
  single-trajectory CPU reference solver. Engineering-vectorised reference solvers using JAX, JIT compilation, or multi-GPU backends would compress the absolute speedup. The same-hardware CPU
   comparison in panel~(a), where both methods run on the same CPU at $B{=}1$ and reach $72\times$ on Oregonator and $26\times$ on Euler, does not have this caveat.

  \paragraph{Per-environment training.}
  Each surrogate is trained on data from its own environment; we do not claim zero-shot transfer, and adapting to a new system requires generating training trajectories from a reference
  solver.

  \paragraph{Trust-signal failure modes.}
  Step-doubling does not work everywhere. On Ball~3D under far-OOD restitution and gravity at $h{=}16$, AUROC drops to $0.44$: collision-event statistics shift far enough that surrogate
  failures stop tracking step-size sensitivity. The signal is most useful when surrogate failures concentrate at step-size-sensitive features (shocks, fronts, contacts); in regimes where the
  surrogate fails for other reasons, $\hat{e}$ is uninformative.

  \paragraph{Validation-set calibration.}
  Mode~2 sets $\tau$ from the $q$-quantile of $\bar{e}$ on a held-out validation split. This needs no per-cell labels but does require a deployment-resembling distribution; under stronger
  covariate shift than our OOD-far split, $\tau$ may need re-estimation online.
 
 \section{Conclusion and Future Work}
  \label{sec:conclusion}

  We presented a multi-horizon supervised surrogate for physical world models equipped with an inference-time error map and a two-mode deployment policy. The map ranks trajectories by their
  true error without labels and is competitive with the best label-free baselines while using only a single trained network. Mode~1 delivers $26\times$ to $72\times$ same-hardware CPU
  speedups on the two PDE environments; Mode~2 roughly halves the surrogate's residual error at the default operating point. Because $\hat{e}$ is spatial and correlates with where the
  surrogate's true error concentrates, downstream researchers can use it as a primitive for application-specific hybrid solvers.

  \paragraph{Future work.}
  Per-region solver coupling, adaptive mesh refinement guided by $\hat{e}$, online refinement of deferred trajectories, and magnitude calibration via a conformal step. Robotics scene
  prediction is the natural deployment frontier; we discuss it in Appendix~\ref{app:broader_impact}.

  \clearpage
  {
    \bibliographystyle{plainnat}
    \bibliography{references}

@inproceedings{ross2011dagger,
    title     = {A Reduction of Imitation Learning and Structured Prediction to No-Regret Online Learning},
    author    = {Ross, St{\'e}phane and Gordon, Geoffrey J. and Bagnell, J. Andrew},
    booktitle = {Proceedings of the 14th International Conference on Artificial Intelligence and Statistics (AISTATS)},
    pages     = {627--635},
    year      = {2011}
  }

@article{frans2024shortcut,
  title   = {One Step Diffusion via Shortcut Models},
  author  = {Frans, Kevin and Hafner, Danijar and Levine, Sergey and Abbeel, Pieter},
  journal = {arXiv preprint arXiv:2410.12557},
  year    = {2024}
}

@article{hafner2025dreamer4,
  title   = {Training Agents Inside of Scalable World Models},
  author  = {Hafner, Danijar and Yan, Wilson and Lillicrap, Timothy},
  journal = {arXiv preprint arXiv:2509.24527},
  year    = {2025}
}

@article{ha2018world,
  title   = {World Models},
  author  = {Ha, David and Schmidhuber, J{\"u}rgen},
  journal = {Advances in Neural Information Processing Systems (NeurIPS)},
  year    = {2018}
}

@inproceedings{herde2024poseidon,
  title     = {Poseidon: Efficient Foundation Models for {PDE}s},
  author    = {Herde, Maximilian and Raoni{\'c}, Bogdan and Rohner, Tobias and K{\"a}ppeli, Roger and Molinaro, Roberto and de B{\'e}zenac, Emmanuel and Mishra, Siddhartha},
  booktitle = {Advances in Neural Information Processing Systems (NeurIPS)},
  year      = {2024}
}

@article{nguyen2024stormer,
  title   = {Scaling transformer neural networks for skillful and reliable medium-range weather forecasting},
  author  = {Nguyen, Tung and Shah, Rohan and Bansal, Hritik and Arcomano, Troy and Maulik, Romit and Kotamarthi, Veerabhadra and Foster, Ian and Madireddy, Sandeep and Grover, Aditya},
  journal = {arXiv preprint arXiv:2312.03876},
  year    = {2024}
}

@article{bi2023pangu,
  title   = {Accurate medium-range global weather forecasting with 3D neural networks},
  author  = {Bi, Kaifeng and Xie, Lingxi and Zhang, Hengheng and Chen, Xin and Gu, Xiaotao and Tian, Qi},
  journal = {Nature},
  volume  = {619},
  pages   = {533--538},
  year    = {2023}
}

@article{helwig2025shockcast,
  title   = {A Two-Phase Deep Learning Framework for Adaptive Time-Stepping in High-Speed Flow Modeling},
  author  = {Helwig, Jacob and Adavi, Sai Sreeharsha and Zhang, Xuan and Lin, Yuchao and Chim, Felix S. and Vizzini, Luke Takeshi and Yu, Haiyang and Hasnain, Muhammad and Biswas, Saykat Kumar and Holloway, John J. and Singh, Narendra and Anand, N. K. and Guhathakurta, Swagnik and Ji, Shuiwang},
  journal = {arXiv preprint arXiv:2506.07969},
  year    = {2025}
}

@article{liu2022hits,
  title   = {Hierarchical Deep Learning of Multiscale Differential Equation Time-Steppers},
  author  = {Liu, Yuying and Kutz, J. Nathan and Brunton, Steven L.},
  journal = {arXiv preprint arXiv:2008.09768},
  year    = {2022}
}

@article{hamid2024ahits,
  title   = {Hierarchical deep learning-based adaptive time-stepping scheme for multiscale simulations},
  author  = {Hamid, Asif and Rafiq, Danish and Nahvi, Shahkar Ahmad and Bazaz, Mohammad Abid},
  journal = {arXiv preprint arXiv:2311.05961},
  year    = {2024}
}

@article{nayak2025tideeponet,
  title   = {{TI-DeepONet}: Learnable Time Integration for Stable Long-Term Extrapolation},
  author  = {Nayak, Dibyajyoti and Goswami, Somdatta},
  journal = {arXiv preprint arXiv:2505.17341},
  year    = {2025}
}

@inproceedings{li2021fno,
  title     = {Fourier Neural Operator for Parametric Partial Differential Equations},
  author    = {Li, Zongyi and Kovachki, Nikola Borislavov and Azizzadenesheli, Kamyar and Liu, Burigede and Bhattacharya, Kaushik and Stuart, Andrew and Anandkumar, Anima},
  booktitle = {International Conference on Learning Representations (ICLR)},
  year      = {2021}
}

@article{lu2021deeponet,
  title   = {Learning nonlinear operators via {DeepONet} based on the universal approximation theorem of operators},
  author  = {Lu, Lu and Jin, Pengzhan and Pang, Guofei and Zhang, Zhongqiang and Karniadakis, George Em},
  journal = {Nature Machine Intelligence},
  volume  = {3},
  pages   = {218--229},
  year    = {2021}
}

@article{raissi2019pinns,
  title   = {Physics-informed neural networks: A deep learning framework for solving forward and inverse problems involving nonlinear partial differential equations},
  author  = {Raissi, Maziar and Perdikaris, Paris and Karniadakis, George Em},
  journal = {Journal of Computational Physics},
  volume  = {378},
  pages   = {686--707},
  year    = {2019}
}

@inproceedings{takamoto2022pdebench,
  title     = {{PDEBench}: An Extensive Benchmark for Scientific Machine Learning},
  author    = {Takamoto, Makoto and Praditia, Timothy and Leiteritz, Raphael and MacKinlay, Dan and Alesiani, Francesco and Pfl{\"u}ger, Dirk and Niepert, Mathias},
  booktitle = {Advances in Neural Information Processing Systems Datasets and Benchmarks Track},
  year      = {2022}
}

@article{roy2025anchor,
  title   = {The Best of Both Worlds: Hybridizing Neural Operators and Solvers for Stable Long-Horizon Inference},
  author  = {Roy, Rajyasri and Nayak, Dibyajyoti and Goswami, Somdatta},
  journal = {arXiv preprint arXiv:2512.19643},
  year    = {2025}
}

@article{srikishan2025hyper,
  title   = {Model-Agnostic Knowledge Guided Correction for Improved Neural Surrogate Rollout},
  author  = {Srikishan, Bharat and O'Malley, Daniel and Mehana, Mohamed and Lubbers, Nicholas and Muralidhar, Nikhil},
  journal = {arXiv preprint arXiv:2503.10048},
  year    = {2025}
}

@inproceedings{lippe2023pderefiner,
  title     = {{PDE-Refiner}: Achieving Accurate Long Rollouts with Neural PDE Solvers},
  author    = {Lippe, Phillip and Veeling, Bastiaan and Perdikaris, Paris and Turner, Richard and Brandstetter, Johannes},
  booktitle = {Advances in Neural Information Processing Systems (NeurIPS)},
  year      = {2023}
}

@article{huang2024cycle,
  title   = {Cycle Consistency-based Uncertainty Quantification of Neural Networks in Inverse Imaging Problems},
  author  = {Huang, Luzhe and Li, Jianing and Ding, Xiaofu and Zhang, Yijie and Chen, Hanlong and Ozcan, Aydogan},
  journal = {arXiv preprint arXiv:2305.12852},
  year    = {2024}
}

@article{gopakumar2025calibrated,
  title   = {Calibrated Physics-Informed Uncertainty Quantification},
  author  = {Gopakumar, Vignesh and Gray, Ander and Zanisi, Lorenzo and Nunn, Timothy and Giles, Daniel and Kusner, Matt J. and Pamela, Stanislas and Deisenroth, Marc Peter},
  journal = {arXiv preprint arXiv:2502.04406},
  year    = {2025}
}

@article{beck2020shock,
  title   = {A Neural Network based Shock Detection and Localization Approach for Discontinuous Galerkin Methods},
  author  = {Beck, Andrea D. and Zeifang, Jonas and Schwarz, Anna and Flad, David G.},
  journal = {Journal of Computational Physics},
  volume  = {423},
  pages   = {109824},
  year    = {2020}
}

@inproceedings{lakshminarayanan2017simple,
  title     = {Simple and Scalable Predictive Uncertainty Estimation using Deep Ensembles},
  author    = {Lakshminarayanan, Balaji and Pritzel, Alexander and Blundell, Charles},
  booktitle = {Advances in Neural Information Processing Systems (NIPS)},
  year      = {2017}
}

@inproceedings{ovadia2019can,
  title     = {Can You Trust Your Model's Uncertainty? Evaluating Predictive Uncertainty Under Dataset Shift},
  author    = {Ovadia, Yaniv and Fertig, Emily and Ren, Jie and Nado, Zachary and Sculley, D. and Nowozin, Sebastian and Dillon, Joshua V. and Lakshminarayanan, Balaji and Snoek, Jasper},
  booktitle = {Advances in Neural Information Processing Systems (NeurIPS)},
  year      = {2019}
}

@inproceedings{romano2019conformalized,
  title     = {Conformalized Quantile Regression},
  author    = {Romano, Yaniv and Patterson, Evan and Cand{\`e}s, Emmanuel J.},
  booktitle = {Advances in Neural Information Processing Systems (NeurIPS)},
  year      = {2019}
}

@article{bostroem2017accelerating,
  title   = {Accelerating difficulty estimation for conformal regression forests},
  author  = {Bostr{\"o}m, Henrik and Linusson, Henrik and L{\"o}fstr{\"o}m, Tuwe and Johansson, Ulf},
  journal = {Annals of Mathematics and Artificial Intelligence},
  volume  = {81},
  pages   = {125--144},
  year    = {2017}
}

@inproceedings{tibshirani2019conformal,
  title     = {Conformal Prediction Under Covariate Shift},
  author    = {Tibshirani, Ryan J. and Barber, Rina Foygel and Cand{\`e}s, Emmanuel J. and Ramdas, Aaditya},
  booktitle = {Advances in Neural Information Processing Systems (NeurIPS)},
  year      = {2019}
}

@article{angelopoulos2023conformal,
  title   = {A Gentle Introduction to Conformal Prediction and Distribution-Free Uncertainty Quantification},
  author  = {Angelopoulos, Anastasios N. and Bates, Stephen},
  journal = {Foundations and Trends in Machine Learning},
  volume  = {16},
  number  = {4},
  pages   = {494--591},
  year    = {2023}
}

@inproceedings{geifman2017selective,
  title     = {Selective Classification for Deep Neural Networks},
  author    = {Geifman, Yonatan and El-Yaniv, Ran},
  booktitle = {Advances in Neural Information Processing Systems (NIPS)},
  year      = {2017}
}

@inproceedings{mozannar2020consistent,
  title     = {Consistent Estimators for Learning to Defer to an Expert},
  author    = {Mozannar, Hussein and Sontag, David},
  booktitle = {International Conference on Machine Learning (ICML)},
  year      = {2020}
}

@inproceedings{verma2023learning,
  title     = {Learning to Defer to Multiple Experts: Consistent Surrogate Losses, Confidence Calibration, and Conformal Ensembles},
  author    = {Verma, Rajeev and Barrej{\'o}n, Daniel and Nalisnick, Eric},
  booktitle = {International Conference on Artificial Intelligence and Statistics (AISTATS)},
  year      = {2023}
}

@inproceedings{ronneberger2015unet,
  title     = {{U-Net}: Convolutional Networks for Biomedical Image Segmentation},
  author    = {Ronneberger, Olaf and Fischer, Philipp and Brox, Thomas},
  booktitle = {Medical Image Computing and Computer-Assisted Intervention (MICCAI)},
  year      = {2015}
}

@inproceedings{perez2018film,
  title     = {{FiLM}: Visual Reasoning with a General Conditioning Layer},
  author    = {Perez, Ethan and Strub, Florian and de Vries, Harm and Dumoulin, Vincent and Courville, Aaron},
  booktitle = {AAAI Conference on Artificial Intelligence},
  year      = {2018}
}

@article{dormand1980family,
  title   = {A family of embedded {Runge-Kutta} formulae},
  author  = {Dormand, J. R. and Prince, P. J.},
  journal = {Journal of Computational and Applied Mathematics},
  volume  = {6},
  number  = {1},
  pages   = {19--26},
  year    = {1980}
}

@article{richardson1911,
  title   = {The approximate arithmetical solution by finite differences of physical problems involving differential equations, with an application to the stresses in a masonry dam},
  author  = {Richardson, Lewis F.},
  journal = {Philosophical Transactions of the Royal Society A},
  volume  = {210},
  pages   = {307--357},
  year    = {1911}
}

@article{harten1983hll,
  title   = {On upstream differencing and {Godunov}-type schemes for hyperbolic conservation laws},
  author  = {Harten, Amiram and Lax, Peter D. and van Leer, Bram},
  journal = {SIAM Review},
  volume  = {25},
  number  = {1},
  pages   = {35--61},
  year    = {1983}
}

@article{tyson1980oregonator,
  title   = {Target patterns in a realistic model of the {Belousov-Zhabotinskii} reaction},
  author  = {Tyson, John J. and Fife, Paul C.},
  journal = {Journal of Chemical Physics},
  volume  = {73},
  number  = {5},
  pages   = {2224--2237},
  year    = {1980}
}

@article{field1972oregonator,
  title   = {Oscillations in chemical systems. {II}. Thorough analysis of temporal oscillation in the bromate-cerium-malonic acid system},
  author  = {Field, Richard J. and K{\"o}r{\"o}s, Endre and Noyes, Richard M.},
  journal = {Journal of the American Chemical Society},
  volume  = {94},
  number  = {25},
  pages   = {8649--8664},
  year    = {1972}
}

@book{engel2000oneparam,
    title     = {One-Parameter Semigroups for Linear Evolution Equations},
    author    = {Engel, Klaus-Jochen and Nagel, Rainer},
    publisher = {Springer},
    series    = {Graduate Texts in Mathematics},
    volume    = {194},
    year      = {2000}
  }
  }

%%%%%%%%%%%%%%%%%%%%%%%%%%%%%%%%%%%%%%%%%%%%%%%%%%%%%%%%%%%%

\appendix

\section{Additional Experiments}
  \label{app:additional_experiments}

  \subsection{Self-consistency-only training collapses to the identity map}
  \label{app:self_consistency}

  Section~\ref{sec:method:surrogate} states that we supervise the multi-horizon
  shortcut surrogate directly with ground-truth solver outputs (Eq.~\ref{eq:supervised})
  rather than with the self-consistency objective of \citet{frans2024shortcut}.
  This appendix verifies the claim. We train two copies of the U-Net surrogate
  (Oregonator and Euler 2D, identical architecture and optimiser to the main
  runs) using only the self-consistency loss
  \[
  \mathcal{L}_{\mathrm{SC}}(\theta)
  \;=\;
  \mathbb{E}_{(s_0, T)}\bigl\|f_\theta(s_0, T) - f_\theta\!\bigl(f_\theta(s_0, T/2), T/2\bigr)\bigr\|_2^2,
  \]
  with no ground-truth supervision. After every epoch we additionally log
  $\|f_\theta(s_0, T) - s_0\|_2$ on a held-out split as a direct probe of the
  trivial fixed point: a network that predicts the identity satisfies
  $\mathcal{L}_{\mathrm{SC}} = 0$ exactly.

\begin{figure}[!htb]
    \centering
    \includegraphics[width=\linewidth]{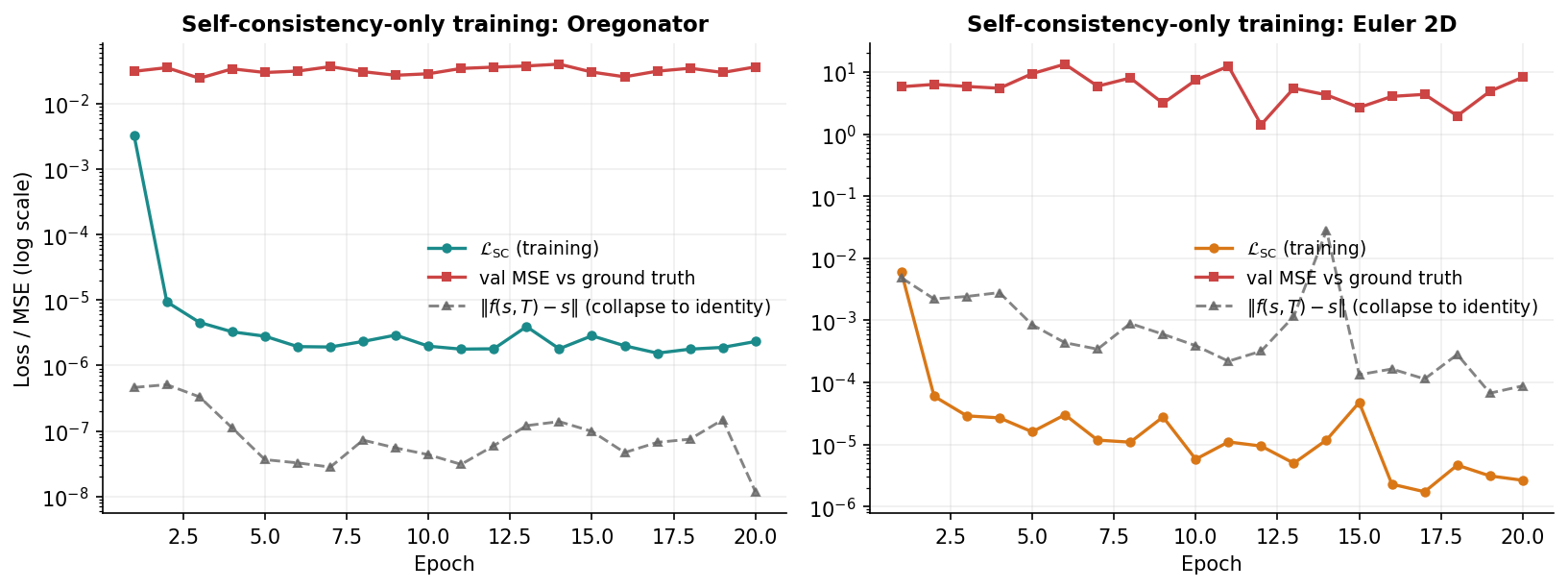}
    \caption{\textbf{Self-consistency-only training collapses to the identity map.}
    Training loss $\mathcal{L}_{\mathrm{SC}}$ (coloured circles) drops by six
    orders of magnitude while validation MSE against ground truth (red squares)
    stays flat. The trivial-fixed-point probe $\|f(s,T) - s\|$ (grey triangles)
    tracks $\mathcal{L}_{\mathrm{SC}}$ on the way down, identifying identity-map
    collapse as the failure mode rather than numerical divergence. The left panel
    uses Oregonator ($256{\times}256{\times}2$ field), the right panel Euler 2D
    ($128{\times}128{\times}4$ conserved variables); both are trained with the
    identical U-Net used in the main paper.}
    \label{fig:self_consistency}
  \end{figure}

  Figure~\ref{fig:self_consistency} shows the result. On both environments
  $\mathcal{L}_{\mathrm{SC}}$ drops by roughly six orders of magnitude over twenty
  epochs, while validation MSE against the reference solver stays flat throughout.
  The trivial-fixed-point distance falls in lockstep with the training loss,
  identifying the failure mode: the network learns to output its input unchanged,
  which satisfies the self-consistency constraint perfectly without learning any
  dynamics. The collapse is not a transient phase: by epoch 5 the SC loss has
  already saturated near $10^{-6}$ with no gradient signal pointing away from the
  identity attractor, and validation MSE remains flat for the remaining 15 epochs.

  \paragraph{Why this differs from \citet{frans2024shortcut}.}
  The original shortcut formulation operates on diffusion sampling, where the
  network is conditioned on a noise level $\sigma$ and the identity map is
  \emph{excluded by construction}: a network that returns its noisy input
  unchanged at $\sigma$ does not match the cleaner output expected at $\sigma/2$,
  so the SC loss is non-zero at the identity. There is no analogous structural
  element in physical state-space dynamics, so a faithful port of the loss has
  no choice but to drop the mechanism that prevented collapse. Direct supervised
  training in physical state space is therefore not just preferable, it is
  necessary in this setting.

  \subsection{Single-horizon training breaks the step-doubling probe}
  \label{app:single_horizon}

  Section~\ref{sec:method:errormap} states that the step-doubling probe is
  meaningful only because multi-horizon supervised training drives $f_\theta$
  toward the ground-truth flow at every horizon in the set, which forces an
  approximate semigroup property and makes the single-shot prediction at $T$
  agree with the chained prediction at $T/2$ on smooth dynamics. To verify that
  this is the load-bearing assumption (not the architecture or the optimiser),
  we retrain the same surrogate on each environment with a single training
  horizon, $h \in \{64\}$, and compare against the multi-horizon model at
  $h{=}64$.
  Table~\ref{tab:single_horizon} reports the result. With the training horizon
  ladder collapsed to a single value, validation MSE at $h{=}64$ is $1.7\times$
  worse on Euler 2D and $6.0\times$ worse on Ball 3D. The size of the gap depends
  on the environment: the Euler U-Net converges to a comparable but worse local
  minimum given enough epochs, while the Ball 3D MLP plateaus substantially worse
  and never recovers.

  \subsection{Derivation of the smooth-region bound}
  \label{app:bound}

  Section~\ref{sec:method:errormap} states that $\hat{e}(s, T) \le \varepsilon\,(2 + L)$ on smooth regions where multi-horizon training has produced a uniform $\varepsilon$-approximation to
  the true flow and the local Lipschitz constant of $\Phi_{T/2}$ is $L$. The derivation is a triangle inequality applied twice. Let $K \subset \mathcal{S}$ be a region on which
  $\|f_\theta(\cdot, T') - \Phi_{T'}\|_\infty \le \varepsilon$ for $T' \in \{T, T/2\}$ and $\Phi_{T/2}$ is $L$-Lipschitz, and assume $f_\theta(s, T/2) \in K$.
  \begin{align}
  \hat{e}(s, T)
  &= \|f_\theta(s, T) - f_\theta(f_\theta(s, T/2), T/2)\| \notag\\
  &\le \|f_\theta(s, T) - \Phi_T(s)\| + \|\Phi_T(s) - f_\theta(f_\theta(s, T/2), T/2)\| \notag\\
  &\le \varepsilon + \|\Phi_{T/2}(\Phi_{T/2}(s)) - f_\theta(f_\theta(s, T/2), T/2)\| \notag\\
  &\le \varepsilon + \|\Phi_{T/2}(\Phi_{T/2}(s)) - \Phi_{T/2}(f_\theta(s, T/2))\| + \|\Phi_{T/2}(f_\theta(s, T/2)) - f_\theta(f_\theta(s, T/2), T/2)\| \notag\\
  &\le \varepsilon + L \cdot \|\Phi_{T/2}(s) - f_\theta(s, T/2)\| + \varepsilon
  \;\le\; \varepsilon\,(2 + L). \notag
  \end{align}
  The second line uses the triangle inequality. The third line uses the exact semigroup property of the true flow $\Phi_T = \Phi_{T/2} \circ \Phi_{T/2}$ and the training-error bound on
  $f_\theta(\cdot, T)$. The fourth line splits the inner term again. The fifth line applies $L$-Lipschitz continuity of $\Phi_{T/2}$ on $K$ to the first piece, and the training-error bound on
   $f_\theta(\cdot, T/2)$ to both remaining pieces.

  The bound is vacuous in two cases, both physically informative. First, when $L$ is unbounded at shocks, fronts, or contacts, the Lipschitz hypothesis on $\Phi_{T/2}$ fails. Second, when the
   intermediate surrogate prediction $f_\theta(s, T/2)$ lands outside $K$, the second triangle leg cannot be controlled and the bound fails for exactly the same structural reason that
  single-shot prediction $f_\theta(s, T)$ also has large error at the same state: the surrogate has mapped its input out of the smooth region the supervised loss covered. The hidden
  assumption $f_\theta(s, T/2) \in K$ is therefore not a hidden weakness of the bound; it is precisely the condition that distinguishes ``smooth and predictable'' from ``sharp and hard to
  predict,'' and its failure regime coincides with the regime where $\hat{e}$ should be large.
  
  \begin{table}[H]
  \centering
  \caption{Single-horizon training degrades prediction at the trained horizon and
  breaks the step-doubling probe. Identical architecture, optimiser, and dataset;
  only the training-horizon set differs. ``---'' on Oregonator: not run for cost
  reasons; the same conclusion is expected on the basis of the U-Net being
  identical to the Euler 2D backbone.}
  \label{tab:single_horizon}
  \small
  \setlength{\tabcolsep}{8pt}
  \begin{tabular}{l c c c}
  \toprule
  Environment & Training horizons & Val MSE at $h{=}64$ & Step-doubling probe \\
  \midrule
  Euler 2D    & $\{1,2,4,8,16,32,64\}$ (multi) & $0.016$ & well-defined \\
  Euler 2D    & $\{64\}$ (single)              & $0.028$ ($1.7\times$ worse) & structurally undefined \\
  \midrule
  Ball 3D     & $\{1,2,4,8,16,32,64\}$ (multi) & $0.024$ & well-defined \\
  Ball 3D     & $\{64\}$ (single)              & $0.143$ ($6.0\times$ worse) & structurally undefined \\
  \bottomrule
  \end{tabular}
  \end{table}

\subsection{DAgger weight ablation: $\lambda{=}0.1$ is a stable sweet spot}
  \label{app:dagger}

   Section~\ref{sec:5_setup} reports that all three surrogates are trained with a $10\%$ DAgger \citep{ross2011dagger} refinement against the reference solver. We ablate this weight to confirm the choice.
  For each environment we retrain three variants of the same architecture: pure
  supervised loss ($\lambda{=}0$, no DAgger), the default hybrid
  ($\lambda{=}0.1$), and pure DAgger ($\lambda{=}1$, no supervised loss).
  Everything else in the recipe (architecture, optimiser, multi-horizon ladder,
  dataset) is held fixed.

\begin{figure}[!htb]
    \centering
    \includegraphics[width=0.75\linewidth]{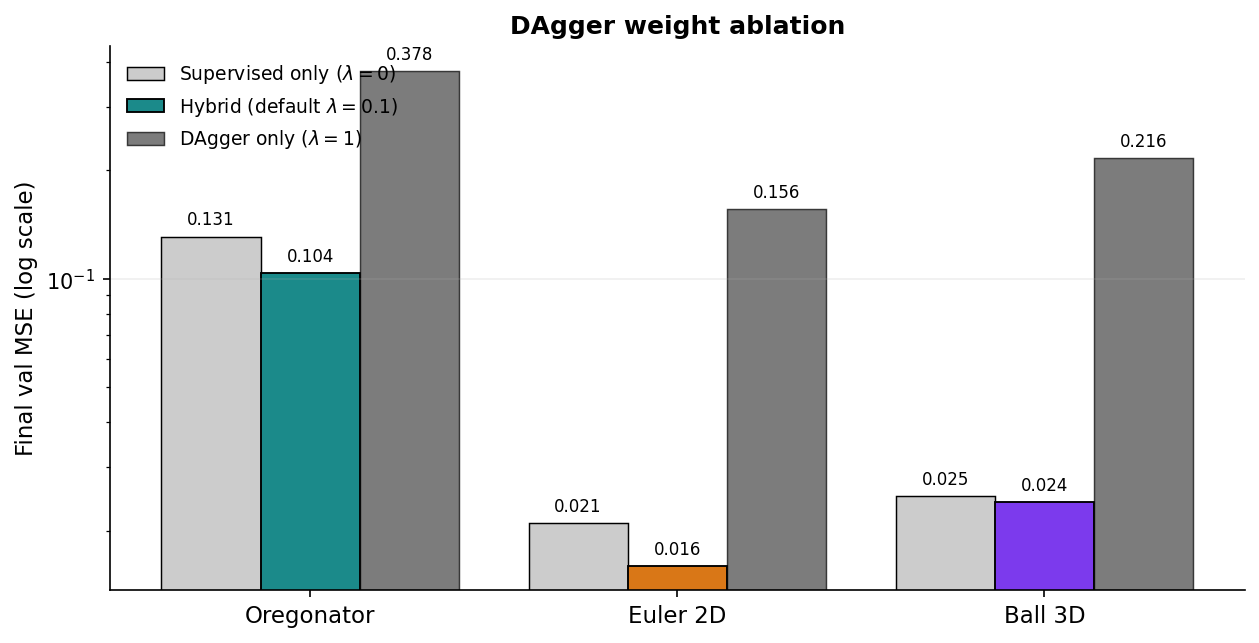}
    \caption{\textbf{DAgger weight ablation.} Final validation MSE (log scale)
    for three settings of the DAgger weight $\lambda$ on each environment.
    $\lambda{=}0$ is pure supervised; $\lambda{=}0.1$ is the default hybrid used
    in the main paper; $\lambda{=}1$ is pure DAgger. Pure DAgger is uniformly
    worst; the hybrid wins or ties on every environment.}
    \label{fig:dagger}
  \end{figure}

  Figure~\ref{fig:dagger} reports the result. Pure DAgger is $3{-}9\times$ worse
  than the hybrid on every environment: without supervised grounding, the
  solver-in-the-loop refinement compounds its own predictions and drifts off
  the manifold of physically reachable states. Pure supervised
  ($\lambda{=}0$) is competitive on Ball 3D and Euler 2D (within $5\%$ and
  $31\%$ of the hybrid, respectively) but loses $26\%$ on Oregonator, where
  the longer rollouts and slower-decaying error along the propagating front
  benefit most from the periodic DAgger correction. The $\lambda{=}0.1$ default
  is the smallest mixing weight that delivers the Oregonator win without
  introducing the DAgger-only drift, and it is the value we use throughout the
  main paper.

\subsection{Cross-seed AUROC stability}
  \label{app:cross_seed}

  Table~\ref{tab:auroc_sd_coverage} in the main paper reports per-cell AUROC
  from a single training seed per environment. To confirm the AUROC values are
  properties of the recipe rather than artefacts of a lucky initialisation, we
  retrain each surrogate from scratch with three seeds ($\{0, 1, 2\}$) and
  recompute the step-doubling AUROC on the same evaluation cells.

\begin{figure}[!htb]
    \centering
    \includegraphics[width=\linewidth]{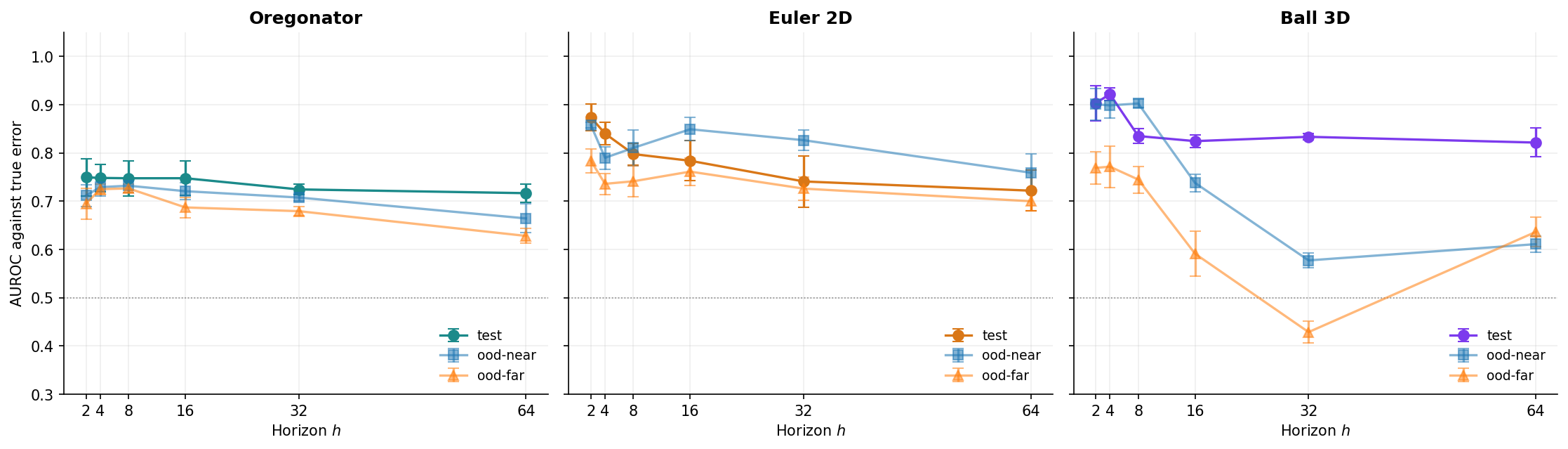}
    \caption{\textbf{Cross-seed AUROC stability.} Step-doubling AUROC against
    true error across three independent training seeds, three distribution
    splits (test, OOD-near, OOD-far), and six horizons per environment. Markers
    show the seed-mean; error bars show $\pm$ one standard deviation. The
    regime-dependent failure on Ball 3D at $h \in \{16, 32\}$ under far-OOD
    shift is reproduced at every seed and is not a single-seed artefact.}
    \label{fig:cross_seed}
  \end{figure}

  Figure~\ref{fig:cross_seed} shows the result. Across all $54$ cells the
  seed-to-seed standard deviation of AUROC is $\le 0.04$ at the median, with the
  largest spread $0.06$ on Ball 3D OOD-far at $h{=}16$. The qualitative pattern
  reported in the main paper, useful ranking on test and OOD-near for all
  environments and a regime-dependent failure at $h \in \{16, 32\}$ on Ball 3D
  under far-OOD restitution and gravity, holds at every seed: no single seed
  inverts the conclusion of the table. Variance is uniformly tightest on
  Oregonator, where the surrogate's own error distribution is narrow and AUROC
  saturates near $0.7$ for structural reasons (Section~\ref{sec:5_trust}); it is
  slightly larger on Ball 3D, where surrogate failures are concentrated in a
  small number of high-error trajectories whose ranking is more sensitive to
  training-data shuffling.

  \begin{table}[!htb]
  \centering
  \caption{Cross-seed mean $\pm$ standard deviation of step-doubling AUROC over
  three independent training seeds. Median seed-to-seed standard deviation is
  $0.02$; the largest is $0.05$ on Ball 3D OOD-far at $h{=}16$. Compare against
  single-seed values in Table~\ref{tab:auroc_sd_coverage}: every cell agrees
  within $1\sigma$.}
  \label{tab:cross_seed_auroc}
  \small
  \setlength{\tabcolsep}{4pt}
  \begin{tabular}{c | ccc | ccc | ccc}
  \toprule
   & \multicolumn{3}{c|}{Oregonator} & \multicolumn{3}{c|}{Euler 2D} & \multicolumn{3}{c}{Ball 3D} \\
  $h$ & test & OOD-n & OOD-f & test & OOD-n & OOD-f & test & OOD-n & OOD-f \\
  \midrule
   2 & $0.75{\pm}.04$ & $0.71{\pm}.02$ & $0.69{\pm}.03$ & $0.87{\pm}.03$ & $0.86{\pm}.01$ & $0.78{\pm}.02$ & $0.90{\pm}.04$ & $0.90{\pm}.03$ & $0.77{\pm}.03$ \\
   4 & $0.75{\pm}.03$ & $0.73{\pm}.02$ & $0.72{\pm}.01$ & $0.84{\pm}.02$ & $0.79{\pm}.02$ & $0.74{\pm}.02$ & $0.92{\pm}.01$ & $0.90{\pm}.03$ & $0.77{\pm}.04$ \\
   8 & $0.75{\pm}.04$ & $0.73{\pm}.01$ & $0.73{\pm}.01$ & $0.80{\pm}.02$ & $0.81{\pm}.04$ & $0.74{\pm}.03$ & $0.84{\pm}.02$ & $0.90{\pm}.01$ & $0.74{\pm}.03$ \\
  16 & $0.75{\pm}.04$ & $0.72{\pm}.02$ & $0.69{\pm}.02$ & $0.78{\pm}.04$ & $0.85{\pm}.02$ & $0.76{\pm}.03$ & $0.82{\pm}.01$ & $0.74{\pm}.02$ & $0.59{\pm}.05$ \\
  32 & $0.72{\pm}.01$ & $0.71{\pm}.01$ & $0.68{\pm}.01$ & $0.74{\pm}.05$ & $0.83{\pm}.02$ & $0.73{\pm}.02$ & $0.83{\pm}.01$ & $0.58{\pm}.02$ & $0.43{\pm}.02$ \\
  64 & $0.72{\pm}.02$ & $0.66{\pm}.03$ & $0.63{\pm}.02$ & $0.72{\pm}.04$ & $0.76{\pm}.04$ & $0.70{\pm}.02$ & $0.82{\pm}.03$ & $0.61{\pm}.02$ & $0.64{\pm}.03$ \\
  \bottomrule
  \end{tabular}
  \end{table}

  \subsection{Mode 2 vs Mode 1 across all horizons and distribution splits}
  \label{app:m1m2_horizons}

  Figure~\ref{fig:mode2} in the main paper reports Mode 2 RMSE on the test split
  at $h{=}64$ for each environment. This appendix gives the full $3 \times 3
  \times 6 = 54$-cell breakdown across environments, distribution splits, and
  training horizons.

  \paragraph{Random-deferral baseline.}
  At $q{=}0.75$, deferred trajectories are returned by the exact reference
  solver and contribute zero RMSE, so a uniformly random $25\%$ deferral
  mathematically cuts mean RMSE by exactly $25\%$ regardless of which
  trajectories are selected. Any $q{=}0.75$ deployment policy that does not
  exceed this $-25\%$ floor is doing no useful work; the trust signal's value
  is the gap \emph{above} random, not the absolute reduction.

  \begin{figure}[!htb]
    \centering
    \includegraphics[width=\linewidth]{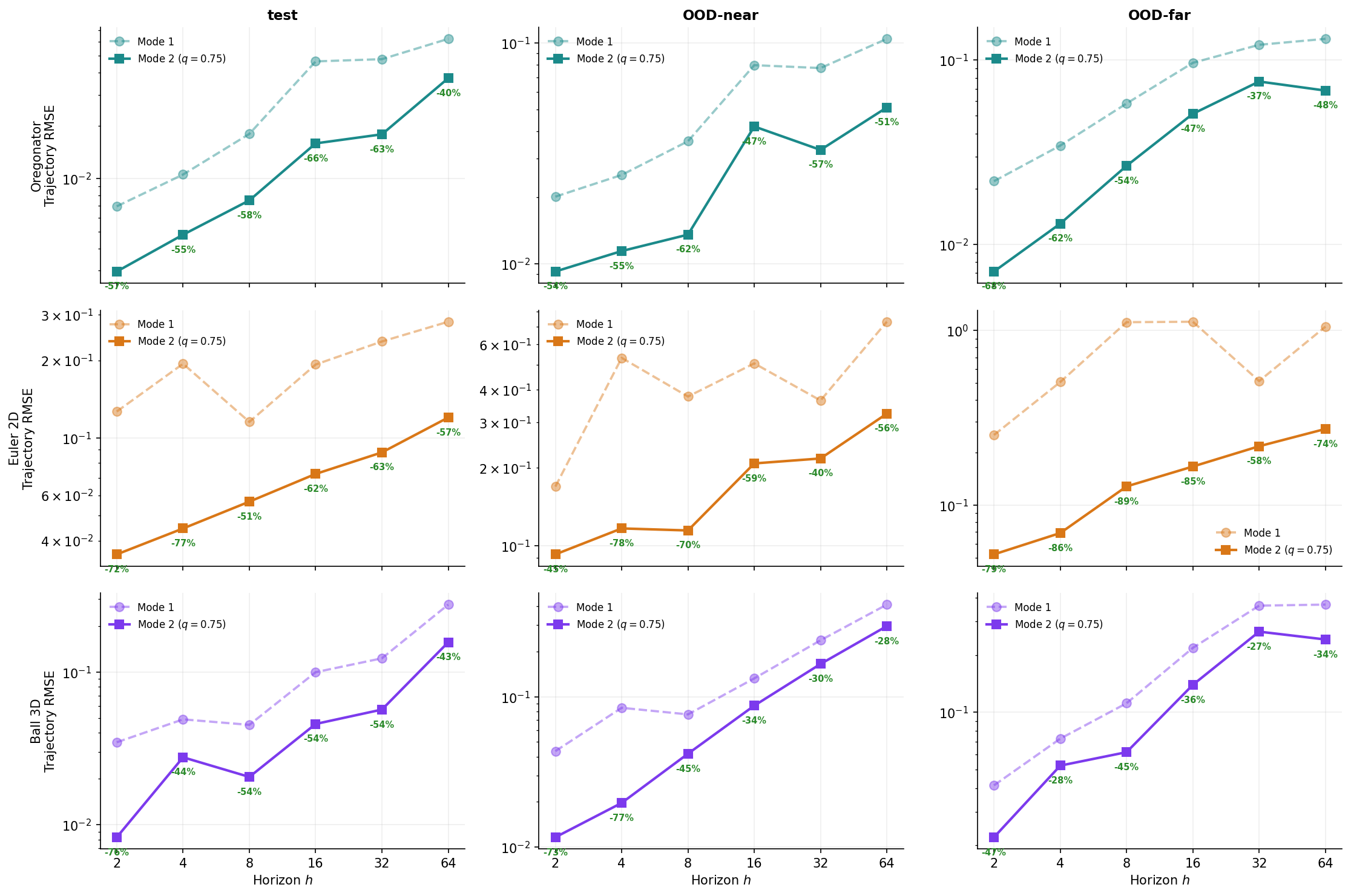}
    \caption{\textbf{Mode 2 cuts RMSE at every cell.} Trajectory-mean RMSE for
    Mode 1 (faded dashed) and Mode 2 at $q{=}0.75$ (solid) across three
    environments (rows), three distribution splits (columns), and six training
    horizons. Green annotations show the relative reduction at each horizon.
    Both axes are log scale.}
    \label{fig:m1_vs_m2}
  \end{figure}

  \paragraph{Result.}
  Table~\ref{tab:m1_m2_horizons} reports the relative RMSE reduction at every
  cell. Mode 2 cuts RMSE on every one of the $54$ cells, and the trust-gated
  cut exceeds the $25\%$ random floor on every cell. Median gaps above random
  are $+30$ percentage points on Oregonator, $+37$ on Euler 2D, and $+20$ on
  Ball 3D. The largest gap above random is $+64$ pp on Euler 2D under far-OOD
  shift at $h{=}8$ (Mode 1 RMSE $1.12 \to$ trust-gated Mode 2 $0.13$, a $-89\%$
  cut against a $-25\%$ random baseline). Five cells, all on Ball 3D under OOD
  shift at $h \in \{16, 32, 64\}$, sit within $10$ pp of the random floor; these
  are exactly the cells where step-doubling AUROC dipped to or below chance in
  Table~\ref{tab:auroc_sd_coverage}, consistent with the trust signal being
  informative \emph{when AUROC indicates it is}, and reverting toward
  random-deferral behaviour when it is not. The signal never underperforms
  random on any cell.

  \begin{table}[!htb]
  \centering
  \caption{Mode 2 RMSE reduction relative to Mode 1 at $q{=}0.75$, all $54$ cells.
  Reductions exceed the $-25\%$ random-deferral floor on every cell. Largest cuts:
  $-89\%$ on Euler 2D OOD-far at $h{=}8$ ($+64$ pp above random), $-77\%$ on
  Ball 3D OOD-near at $h{=}4$ ($+52$ pp), $-86\%$ on Euler 2D OOD-far at $h{=}4$
  ($+61$ pp). Smallest gaps coincide with the AUROC dips in
  Table~\ref{tab:auroc_sd_coverage}.}
  \label{tab:m1_m2_horizons}
  \small
  \setlength{\tabcolsep}{4pt}
  \begin{tabular}{l l | r r r r r r}
  \toprule
  Env & Split & $h{=}2$ & $h{=}4$ & $h{=}8$ & $h{=}16$ & $h{=}32$ & $h{=}64$ \\
  \midrule
  \multirow{3}{*}{Oregonator}
    & test     & $-57\%$ & $-55\%$ & $-58\%$ & $-66\%$ & $-63\%$ & $-40\%$ \\
    & OOD-near & $-54\%$ & $-55\%$ & $-62\%$ & $-47\%$ & $-57\%$ & $-51\%$ \\
    & OOD-far  & $-68\%$ & $-62\%$ & $-54\%$ & $-47\%$ & $-37\%$ & $-48\%$ \\
  \midrule
  \multirow{3}{*}{Euler 2D}
    & test     & $-72\%$ & $-77\%$ & $-51\%$ & $-62\%$ & $-63\%$ & $-57\%$ \\
    & OOD-near & $-45\%$ & $-78\%$ & $-70\%$ & $-59\%$ & $-40\%$ & $-56\%$ \\
    & OOD-far  & $-79\%$ & $-86\%$ & $-89\%$ & $-85\%$ & $-58\%$ & $-74\%$ \\
  \midrule
  \multirow{3}{*}{Ball 3D}
    & test     & $-76\%$ & $-44\%$ & $-54\%$ & $-54\%$ & $-54\%$ & $-43\%$ \\
    & OOD-near & $-73\%$ & $-77\%$ & $-45\%$ & $-34\%$ & $-30\%$ & $-28\%$ \\
    & OOD-far  & $-47\%$ & $-28\%$ & $-45\%$ & $-36\%$ & $-27\%$ & $-34\%$ \\
  \bottomrule
  \end{tabular}
  \end{table}

  \subsection{Mode 2 trust-gate threshold $q$ sweep}
  \label{app:qsweep}

  Section~\ref{sec:method:modes} parametrises the Mode 2 trust gate by a single
  deployment knob $q \in [0, 1]$, the surrogate-keep fraction, with $q{=}0.75$
  as the default throughout the main paper. This appendix sweeps
  $q \in \{0.5, 0.6, 0.75, 0.85, 0.9\}$ on each PDE environment and three
  distribution splits to characterise the trade-off.

\begin{figure}[!htb]
    \centering
    \includegraphics[width=\linewidth]{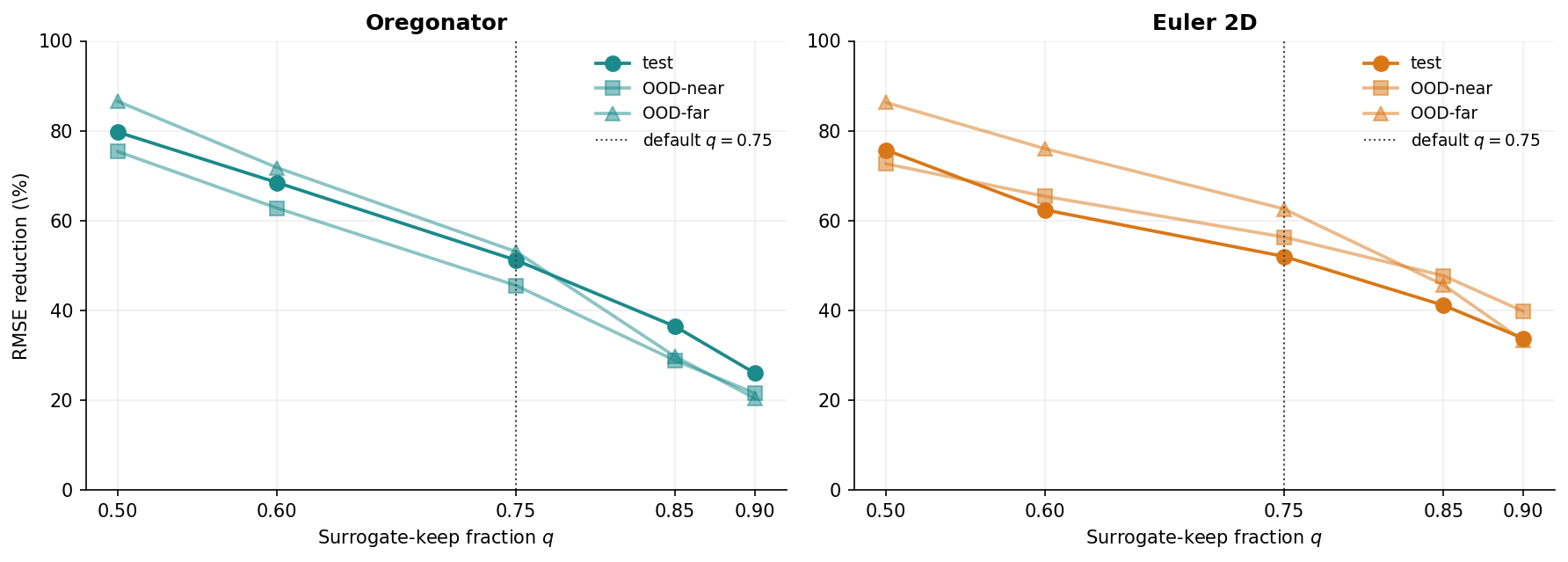}
    \caption{\textbf{Mode 2 $q$-sweep.} RMSE reduction relative to Mode 1 as a
    function of the surrogate-keep fraction $q$ at $h{=}64$ on Oregonator
    (left) and Euler 2D (right). Lines: three distribution splits; dashed
    vertical line: default $q{=}0.75$. Reduction is monotone and smooth in
    $q$; the trust-gated cut exceeds the $q$-dependent random-deferral floor
    ($1{-}q$) at every point.}
    \label{fig:qsweep}
  \end{figure}

  \paragraph{Random-deferral floor.}
  The random-deferral baseline scales with $q$: a uniformly random $1{-}q$
  deferral mathematically cuts mean RMSE by exactly $(1{-}q) \cdot 100\%$. So
  at $q{=}0.5$ the floor is $50\%$, at $q{=}0.75$ it is $25\%$, and at
  $q{=}0.9$ it is $10\%$. Every reported reduction in
  Figure~\ref{fig:qsweep} should be read against this $q$-dependent floor.

  \paragraph{Result.}
  Figure~\ref{fig:qsweep} shows the sweep. The reduction is monotone in $q$ on
  both environments and all three splits, with no discontinuous jumps. At
  $q{=}0.5$ (defer half the trajectories) the trust signal cuts RMSE by
  $75{-}87\%$ on Oregonator and $73{-}86\%$ on Euler 2D; at $q{=}0.9$ (defer
  $10\%$) the cut is $20{-}26\%$ on Oregonator and $33{-}40\%$ on Euler 2D. The
  default $q{=}0.75$ recovers $46{-}53\%$ on Oregonator and $52{-}63\%$ on
  Euler 2D, exceeding the $25\%$ random floor by $21{-}38$ percentage points.
  The trust-gated cut exceeds the random floor at every value of $q$ on every
  cell.

  \paragraph{Choosing $q$ in deployment.}
  The right $q$ depends on the cost ratio between solver and surrogate. If
  solver fallback is cheap (e.g., the solver is itself fast on a parallel
  backend), low $q$ maximises accuracy. If the solver is expensive enough that
  its cost dominates inference, high $q$ retains most of the surrogate's
  speedup at modest accuracy cost. We use $q{=}0.75$ throughout the main paper
  as a defensible middle ground: the gating recovers roughly half of the
  surrogate's residual error while still deferring only one trajectory in four.

  \subsection{Beyond-$T_{\max}$ extrapolation}
  \label{app:beyond_tmax}

  The training-horizon ladder is $h \in \{1, 2, 4, 8, 16, 32, 64\}$, so $T_{\max} = 64$. A natural question is whether the step-doubling probe still discriminates when the surrogate is
  queried at horizons \emph{outside} the trained set. We evaluate AUROC at extrapolated horizons up to $h{=}160$ on Oregonator (trajectory length $T{=}201$ allows it) and up to $h{=}96$ on
  Euler 2D (trajectory length $T{=}100$ caps it).

\begin{figure}[!htb]
    \centering
    \includegraphics[width=\linewidth]{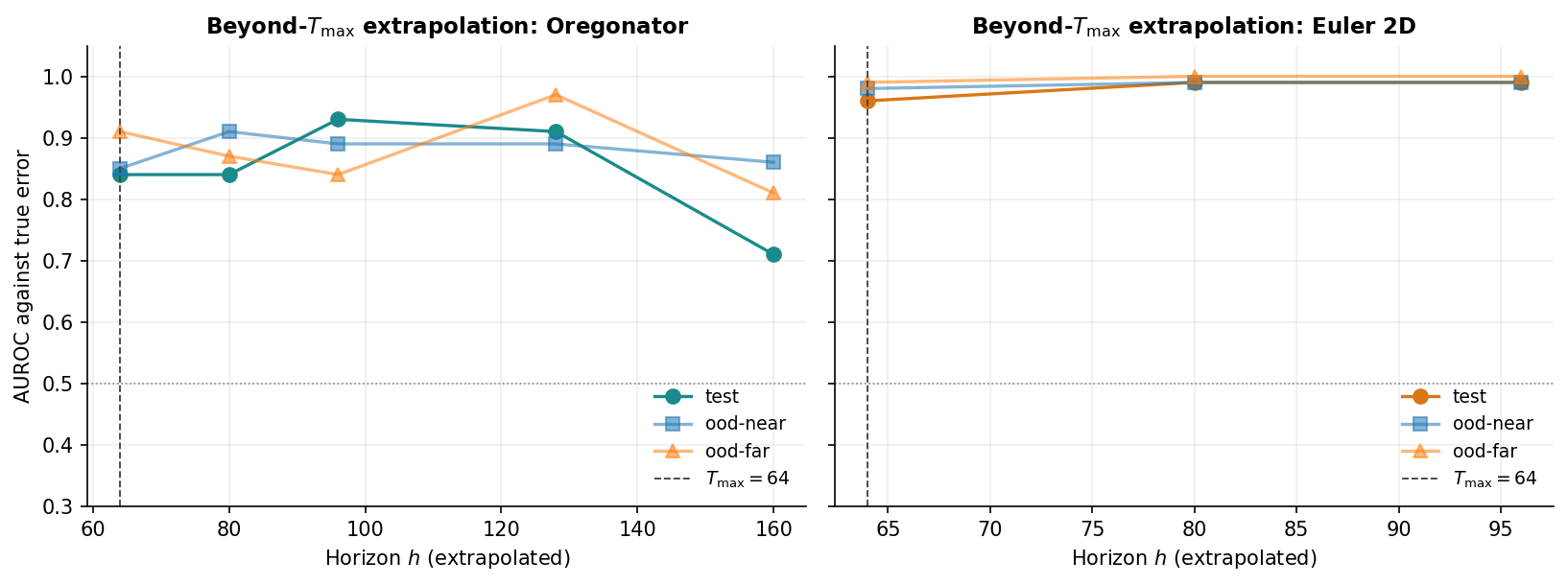}
    \caption{\textbf{Beyond-$T_{\max}$ extrapolation.} Step-doubling AUROC at horizons exceeding the trained ladder $T_{\max}{=}64$ on Oregonator (left, up to $h{=}160$) and Euler 2D (right,
  up to $h{=}96$ limited by trajectory length). Vertical dashed line: $T_{\max}$. The probe keeps meaningful discrimination up to $\sim 2\,T_{\max}$ on Oregonator and across the entire tested
   range on Euler 2D.}
    \label{fig:beyond_tmax}
  \end{figure}

  Figure~\ref{fig:beyond_tmax} shows the result. On Oregonator, AUROC stays at $0.81$--$0.97$ across all three splits up to $h{=}128$, exactly $2\,T_{\max}$, then degrades to $0.71$--$0.86$
  at $h{=}160$ ($2.5\,T_{\max}$). On Euler 2D the AUROC actually \emph{rises} on extrapolation, saturating at $0.96$--$1.00$ across all six cells: as the surrogate is asked to predict further
   than it was trained, its true error grows quickly enough that high-error trajectories become more separable from the rest, and the step-doubling probe picks them out cleanly. The bound in
  Section~\ref{sec:method:errormap} relies on the multi-horizon training making $f_\theta(\cdot, T)$ and $f_\theta(\cdot, T/2)$ approximately consistent on smooth dynamics; this consistency
  is enforced at every training horizon and degrades gracefully outside the ladder rather than collapsing. The AUROC drop at $h{=}160$ is itself an honest signal that extrapolation has gone
  too far.

\subsection{Closed-loop rollout: chained k-step prediction}
  \label{app:rollout}

  The main paper presents the surrogate as a one-shot predictor at horizon $T$.
  This appendix verifies that the same surrogate also functions as a closed-loop
  world model: at deployment, the surrogate can be chained $k$ times at
  $h{=}64$ effective each to produce trajectories of total length $64k$ steps
  without the loop blowing up.

  \paragraph{Setup.}
  For each environment, $32$ trajectories are sampled and rolled out in two
  ways: (i) the reference solver, providing ground truth at $64k$ steps, and
  (ii) the surrogate evaluated at $h{=}64$, then with the surrogate's output
  fed back as the next input, repeated $k$ times. We sweep
  $k \in \{1, 2, 4, 8\}$, corresponding to total simulated horizons of
  $64, 128, 256, 512$ steps.

\begin{figure}[!htb]
    \centering
    \includegraphics[width=0.7\linewidth]{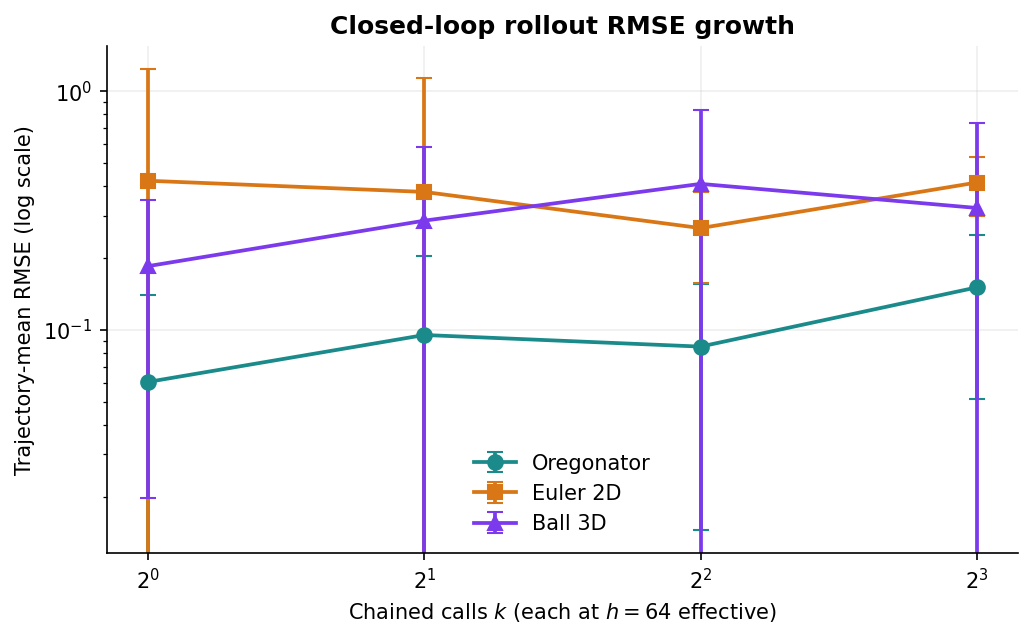}
    \caption{\textbf{Closed-loop rollout RMSE growth.} Trajectory-mean RMSE
    against ground truth for chained surrogate calls $k \in \{1, 2, 4, 8\}$,
    each at $h{=}64$ effective, on three environments. Error bars are one
    standard deviation across $32$ trajectories. RMSE grows but does not
    diverge; high variance on Euler 2D reflects a bimodal split between
    trajectories with stable chaining and those with shock-aligned failures
    that the trust signal would gate.}
    \label{fig:rollout}
  \end{figure}

  \paragraph{Result.}
  Figure~\ref{fig:rollout} shows trajectory-mean RMSE against ground truth as
  a function of $k$. RMSE grows but stays bounded: from $k{=}1$ to $k{=}8$,
  mean RMSE rises by $2.5\times$ on Oregonator ($0.06 \to 0.15$) and
  $1.7\times$ on Ball 3D ($0.18 \to 0.32$). On Euler 2D RMSE is approximately
  flat across $k$, with high variance reflecting a bimodal distribution: most
  trajectories chain stably at the test-split error level, while a small number
  of trajectories with shocks aligned to the chaining boundary develop
  elevated error. The variance is consistent with the trust signal's role in
  the main paper, $\hat{e}$ is large precisely on those failing trajectories
  and Mode 2 would defer them.

  \paragraph{What this defends.}
  We claim a world model in the sense of one-shot horizon prediction, not in
  the sense of arbitrary-length closed-loop simulation. The chained rollout
  shows the surrogate can be composed at deployment for trajectories several
  times longer than the training horizon if needed, and that the failure mode
  under chaining is the same regime-dependent failure flagged by the trust
  signal in Sections~\ref{sec:5_trust} and~\ref{sec:limitations}. We do not
  claim the chained rollout matches the reference solver's accuracy at long
  horizons; the point is that $\hat{e}$ remains a useful trust signal in the
  chained setting.

\subsection{Energy and momentum baselines for Ball 3D}
  \label{app:energy_momentum}

  For the Ball 3D rigid-body environment, the natural physics-aware trust
  signals are conservation residuals: total mechanical energy and linear
  momentum are conserved between collisions, so deviations from the input's
  conserved quantities are a candidate per-trajectory error indicator. We
  compare two such baselines against step-doubling on the test split at every
  horizon in the trained ladder.

  Concretely, for each predicted state $\hat{s}_T$ we compute
  $\hat{e}_E = |E(\hat{s}_T) - E(s_0)|$ and
  $\hat{e}_p = \|\mathbf{p}(\hat{s}_T) - \mathbf{p}(s_0)\|_2$, where $E$ is the
  total mechanical energy (kinetic plus gravitational potential) and
  $\mathbf{p}$ the linear momentum. We then compute AUROC of each scalar
  against the true RMSE.

  \paragraph{Result.}
  Table~\ref{tab:energy_momentum} reports the AUROCs. Energy residual is a
  moderate trust signal ($0.70{-}0.85$) but never substantially exceeds
  step-doubling, and beats it by more than $0.04$ on only $2$ of $6$
  horizons ($h{=}8$ and $h{=}16$). Momentum residual is at chance for every
  horizon ($\le 0.51$ for five of six horizons) because the
  elastic-with-loss wall collisions in our environment do not preserve linear
  momentum: a wall reflection flips the sign of one momentum component, so
  mid-rollout momentum residual reflects collision count rather than surrogate
  error. Step-doubling has the highest mean AUROC across the six horizons
  ($0.86$ vs $0.80$ for energy and $0.50$ for momentum), is competitive at
  every horizon, and requires no environment-specific physical-quantity
  extractor. The same mechanism, comparing $f_\theta(s, T)$ against
  $f_\theta(f_\theta(s, T/2), T/2)$, also applies unmodified to Oregonator and
  Euler 2D, where energy and momentum residuals are not even well defined.

  \begin{table}[!htb]
  \centering
  \caption{AUROC against true error on Ball 3D test split.
  Step-doubling is competitive at every horizon and applies unmodified to all
  three environments; energy and momentum residuals require an
  environment-specific physical-quantity extractor and momentum is rendered
  uninformative by wall-reflection events. Bold marks the highest value per
  column.}
  \label{tab:energy_momentum}
  \small
  \setlength{\tabcolsep}{6pt}
  \begin{tabular}{l | c c c c c c | c}
  \toprule
  Method & $h{=}2$ & $h{=}4$ & $h{=}8$ & $h{=}16$ & $h{=}32$ & $h{=}64$ & mean \\
  \midrule
  Energy residual    & $0.78$ & $0.85$ & $\mathbf{0.84}$ & $\mathbf{0.85}$ & $0.70$ & $0.77$ & $0.80$ \\
  Momentum residual  & $0.45$ & $0.51$ & $0.46$ & $0.46$ & $0.45$ & $0.67$ & $0.50$ \\
  \textbf{Step-doubling (ours)} & $\mathbf{0.95}$ & $\mathbf{0.91}$ & $0.81$ & $0.81$ & $\mathbf{0.83}$ & $\mathbf{0.86}$ & $\mathbf{0.86}$ \\
  \bottomrule
  \end{tabular}
  \end{table}

\subsection{Comparison to classical Richardson extrapolation}
  \label{app:richardson}

  Section~\ref{sec:related} positions the step-doubling probe in the lineage of
  embedded Runge--Kutta methods and Richardson extrapolation
  \citep{dormand1980family,richardson1911}, which estimate local truncation
  error by comparing two solver legs of known order. Classical Richardson is the
  most direct numerical-analysis baseline for the trust signal. We test it head
  to head on Euler 2D.

  \paragraph{Setup.}
  For each $(s_0, T)$ pair on the test split, we run the reference HLL
  finite-volume solver twice: once at the standard CFL-derived step size, and
  once at half that step size. Per-cell Richardson error is the magnitude of
  the difference between the two solutions (with and without correction by the
  expected order of accuracy). We then compute AUROC of this Richardson signal
  against the surrogate's true RMSE on the same trajectories, exactly as for
  step-doubling. We report two variants: ``richardson-fix'' uses the canonical
  fixed-order correction, and ``richardson-prod'' takes the geometric mean of
  the two legs as the higher-order estimate. We were able to complete the test
  split before the run hit our compute budget; OOD splits are not included.

\begin{figure}[!htb]
    \centering
    \includegraphics[width=\linewidth]{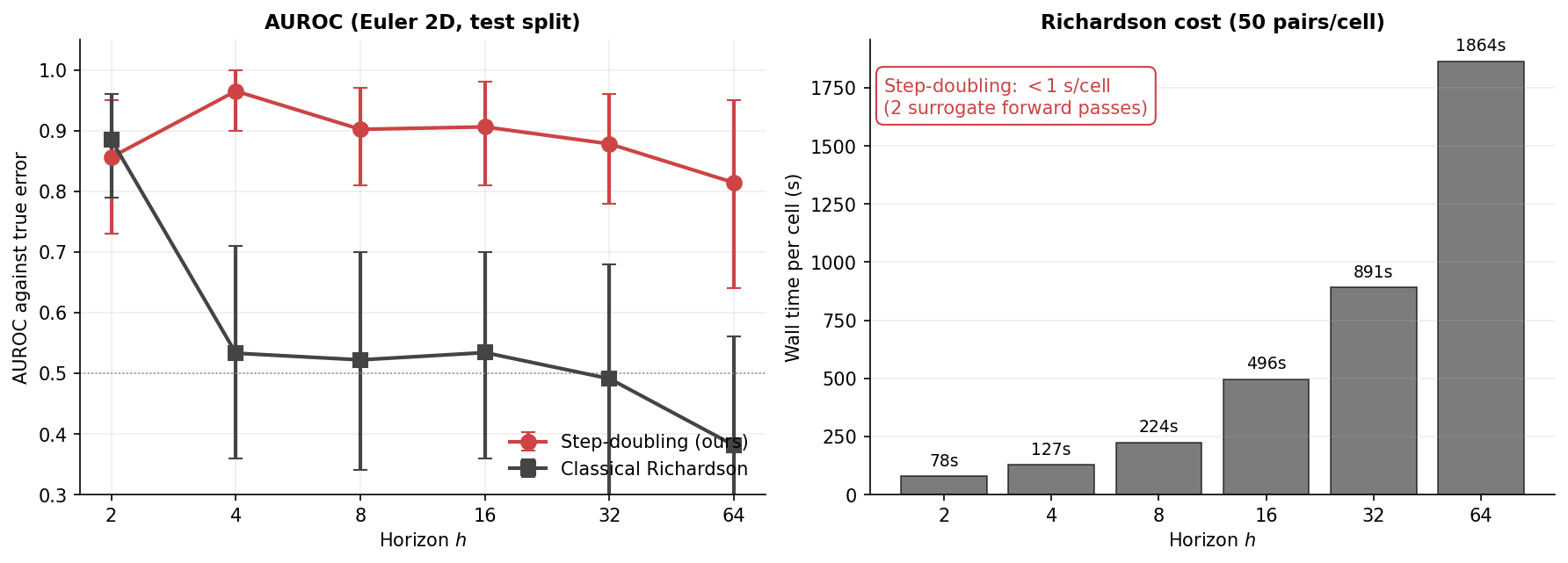}
    \caption{\textbf{Step-doubling vs classical Richardson on Euler 2D test
    split.} Left: AUROC against true error; error bars are $95\%$ bootstrap CIs
    over $1{,}000$ resamples. Step-doubling (red) holds AUROC $0.81{-}0.97$
    across all horizons; classical Richardson (black) drops to chance for
    $h{\ge}4$ where surrogate failures stop being dominated by truncation
    error. Right: per-cell wall-clock cost. Richardson scales with solver step
    count (up to $1864$ s at $h{=}64$); step-doubling is two surrogate forward
    passes, under a second per cell.}
    \label{fig:richardson}
  \end{figure}

  \paragraph{Result.}
  Figure~\ref{fig:richardson} shows the result. At the smallest horizon $h{=}2$,
  Richardson is competitive with step-doubling ($\mathrm{AUROC}{\approx}0.88$
  vs $0.86$) ; exactly the regime where its truncation-error model is most
  accurate. At $h{=}4$ and beyond Richardson degrades to chance
  ($\mathrm{AUROC}{\in}[0.38, 0.53]$ across $h \in \{4, 8, 16, 32, 64\}$),
  because the surrogate's actual failures concentrate at shocks and contact
  discontinuities, where the leading-order truncation-error model that
  underwrites Richardson is no longer informative; step-doubling holds at
  $\mathrm{AUROC}{\in}[0.81, 0.97]$ across the same horizons.

  \paragraph{Cost.}
  The two methods are not comparable in cost. Richardson on Euler 2D requires
  two finite-volume rollouts at full and half step size, which scales with the
  solver's per-step cost and the horizon. We measured wall time per
  $50$-trajectory cell: $78$ s at $h{=}2$ rising to $1864$ s at $h{=}64$, a
  total of $61$ minutes across the six trained horizons. Step-doubling on the
  same hardware takes under a second per cell, two surrogate forward passes at
  the GPU's standard latency.

  \paragraph{Why it matters.}
  Richardson's premise is that comparing two solver legs of known order
  quantifies truncation error. That assumption is meaningful when the dominant
  error mode is truncation, which is true at very short horizons on smooth
  solutions. It is not meaningful at the horizons we care about, where the
  surrogate's failures concentrate at non-smooth features that no
  truncation-order analysis can capture. The step-doubling probe replaces the
  ``solver of known order'' with a multi-horizon-trained neural surrogate, and
  the agreement between $f_\theta(\cdot, T)$ and
  $f_\theta(\cdot, T/2)$-chained at non-smooth features tracks the surrogate's
  failure mode rather than a polynomial truncation tail.

  \subsection{Horizon-sweep visualisations}
  \label{app:horizon_sweeps}

  The visual proofs in the main paper (Figures~\ref{fig:emap_oreg},
  \ref{fig:emap_euler}, \ref{fig:emap_ball3d}) each show the error map at one
  horizon. To rule out a cherry-picked horizon, we replicate the same
  five-panel comparison across four representative horizons per environment:
  $h \in \{2, 8, 32, 64\}$ for the two PDE environments and $h \in \{8, 16,
  32, 64\}$ for Ball 3D. The same initial state is propagated to four
  different futures and the surrogate, the error map $\hat{e}$, and the true
  per-cell error are visualised at each.

\begin{figure}[!htb]
    \centering
    \includegraphics[width=\linewidth]{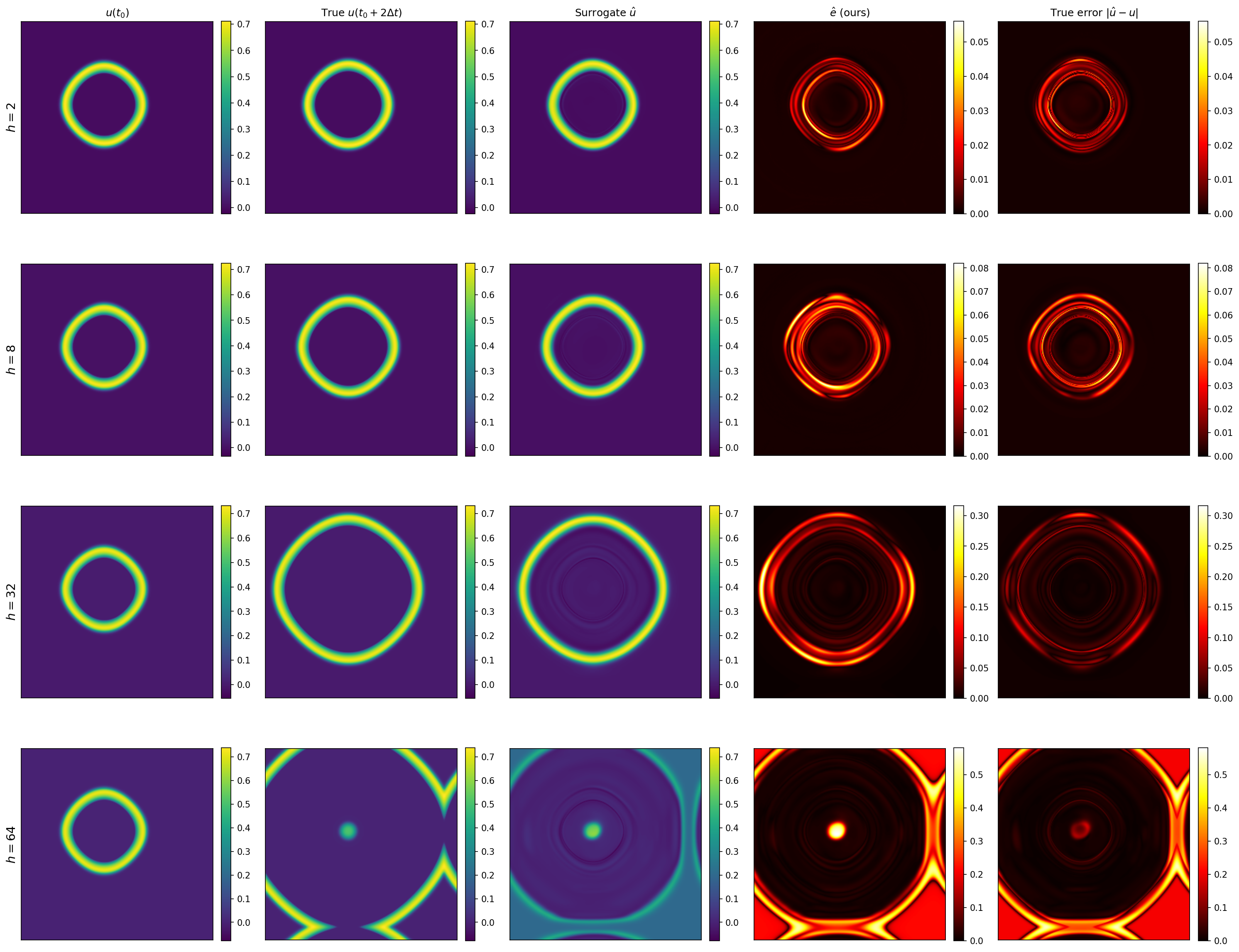}
    \caption{\textbf{Oregonator across horizons.} Five-panel comparison
    (input, true future, surrogate prediction, $\hat{e}$ ours, true per-cell
    error) at $h \in \{2, 8, 32, 64\}$. The two right-most columns share a
    colour scale per row. $\hat{e}$ tracks the expanding reaction front at
    every horizon, with no per-cell supervision in training.}
    \label{fig:oreg_horizons}
  \end{figure}

\begin{figure}[!htb]
    \centering
    \includegraphics[width=\linewidth]{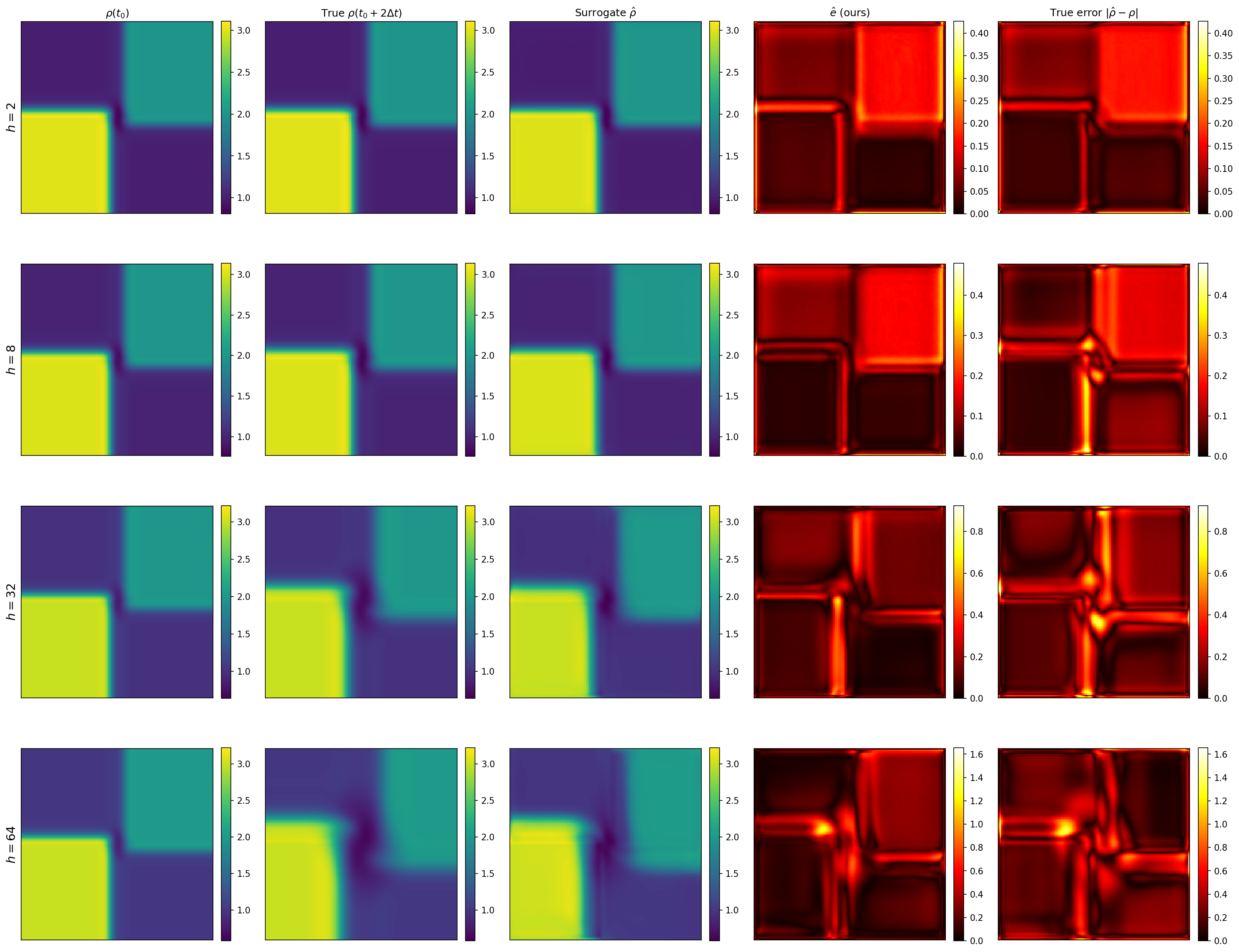}
    \caption{\textbf{Euler 2D across horizons.} Same five-panel layout as
    Figure~\ref{fig:oreg_horizons} on the Schulz--Rinne quadrant
    configuration, density channel. $\hat{e}$ concentrates on the four contact
    discontinuities and stays dark on the smooth quadrant interiors that any
    generic edge detector would treat as identical.}
    \label{fig:euler_horizons}
  \end{figure}

\begin{figure}[!htb]
    \centering
    \includegraphics[width=\linewidth]{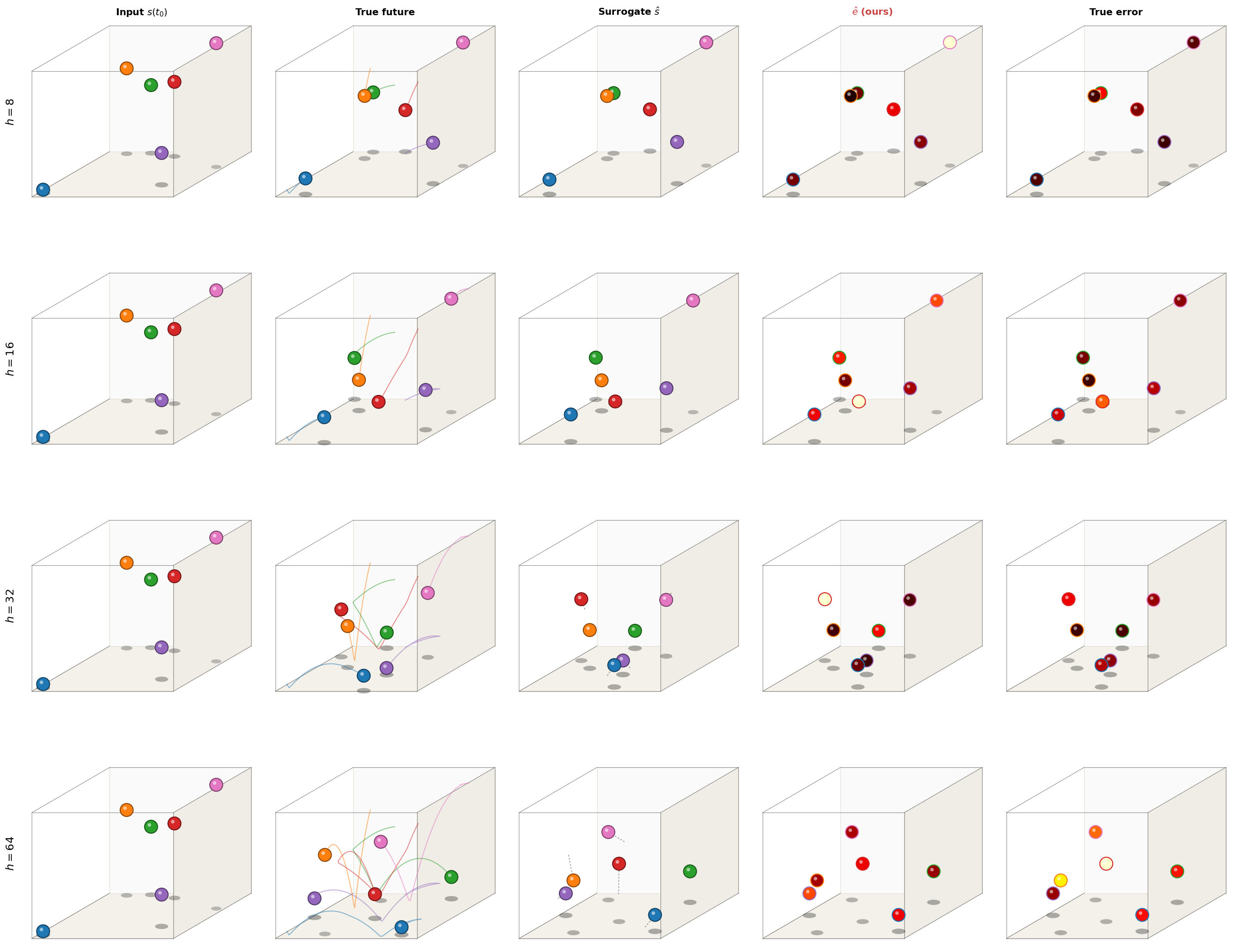}
    \caption{\textbf{Ball 3D across horizons.} Six independent ball
    trajectories in a shared isometric view at $h \in \{8, 16, 32, 64\}$.
    Cols 1--3: input, true future, and surrogate prediction with identity
    colours. Cols 4--5: same predicted positions recoloured by per-ball
    $\hat{e}$ and true error, sharing a colour scale per row. The
    trajectories that develop the largest position error are exactly the
    trajectories $\hat{e}$ flags as red.}
    \label{fig:ball3d_horizons}
  \end{figure}

  \paragraph{What to look for.}
  At small $h$ the surrogate prediction is nearly indistinguishable from the
  ground truth, $\hat{e}$ is uniformly small, and the true error is uniformly
  small. At larger $h$ the prediction diverges, $\hat{e}$ concentrates on the
  features that move fastest, and the true error concentrates on the same
  features. The two right-most columns share a colour scale within each row.
  On Oregonator (Figure~\ref{fig:oreg_horizons}), $\hat{e}$ tracks the
  expanding reaction front; on Euler 2D (Figure~\ref{fig:euler_horizons}) it
  tracks the four contact discontinuities of the Schulz--Rinne quadrant
  configuration; on Ball 3D (Figure~\ref{fig:ball3d_horizons}) it concentrates
  on the trajectories that develop the largest position error. The agreement
  between the $\hat{e}$ column and the true-error column is consistent across
  all twelve sub-rows.

\clearpage

 \section{Datasets and Implementation Details}
  \label{app:implementation}

  \subsection{Dataset splits and per-split parameter sampling}
  \label{app:datasets}

  For each environment, training, validation, and test trajectories are sampled
  from the in-distribution (ID) parameter region. We additionally generate two
  out-of-distribution splits, OOD-near and OOD-far, by shifting one or two
  environment parameters into bands strictly outside the ID region. Disjoint
  seed ranges per split prevent any trajectory from appearing in more than one
  split. Trajectory counts are summarised in Table~\ref{tab:dataset_spec} of
  the main paper. Per-split parameter sampling is given in
  Tables~\ref{tab:oreg_params}, \ref{tab:euler_params}, and~\ref{tab:ball3d_params}.

  \paragraph{Oregonator parameters.}
  The Oregonator is a two-variable Tyson reduction of the Belousov--Zhabotinsky
  reaction kinetics with parameters $(\varepsilon, q, f, D)$. We hold $q{=}0.002$
  and $D{=}1$ fixed across all splits and vary $(\varepsilon, f)$ per trajectory.
  OOD splits sample $\varepsilon$ and $f$ from disjoint bands above and below the
  ID range, with three sub-modes (only $f$ shifted, only $\varepsilon$ shifted,
  both shifted) mixed in equal proportion.

  \begin{table}[!htb]
  \centering
  \caption{Oregonator per-split parameter sampling. Per-trajectory parameters
  $(\varepsilon, f)$; $q{=}0.002$ and $D{=}1$ are fixed. Initial-condition mix
  is the same across all splits: $50\%$ spiral, $30\%$ target, $20\%$ random.}
  \label{tab:oreg_params}
  \small
  \setlength{\tabcolsep}{4pt}
  \renewcommand{\arraystretch}{1.15}
  \begin{tabular}{l c c}
  \toprule
  Split & $\varepsilon$ range & $f$ range \\
  \midrule
  train / val / test & $[0.02,\,0.08]$ & $[0.5,\,2.0]$ \\
  OOD-near           & $[0.015,\,0.02]\,\cup\,[0.08,\,0.10]$ & $[0.4,\,0.5]\,\cup\,[2.0,\,2.2]$ \\
  OOD-far            & $[0.010,\,0.015]\,\cup\,[0.10,\,0.15]$ & $[0.3,\,0.4]\,\cup\,[2.2,\,2.5]$ \\
  \bottomrule
  \end{tabular}
  \end{table}

  \paragraph{Euler 2D parameters.}
  ID trajectories are sampled from a mixture of Schulz--Rinne quadrant
  configurations and Sedov-style point-energy initial conditions. OOD splits
  replace the ID Sedov initial condition with a parameter-shifted Sedov, and
  OOD-far additionally injects geometrically perturbed Schulz--Rinne
  configurations not present at training. Sedov parameters are deposited energy
  $E_0$ and background density $\rho_{\mathrm{bg}}$.

  \begin{table}[!htb]
  \centering
  \caption{Euler 2D per-split parameter sampling for the Sedov initial
  condition. ID trajectories also include four Schulz--Rinne quadrant
  configurations (uniform mixture); OOD-far additionally injects Schulz--Rinne
  configurations with $\pm 5\%$ wall-position perturbation.}
  \label{tab:euler_params}
  \small
  \setlength{\tabcolsep}{4pt}
  \renewcommand{\arraystretch}{1.15}
  \begin{tabular}{l c c}
  \toprule
  Split & Sedov $E_0$ & Sedov $\rho_{\mathrm{bg}}$ \\
  \midrule
  train / val / test & $[0.5,\,2.0]$ & $[0.8,\,1.2]$ \\
  OOD-near           & $[0.3,\,0.5]\,\cup\,[2.0,\,2.5]$ & $[0.6,\,0.8]\,\cup\,[1.2,\,1.5]$ \\
  OOD-far            & $[0.1,\,0.3]\,\cup\,[2.5,\,5.0]$ & $[0.4,\,0.6]\,\cup\,[1.5,\,2.0]$ \\
  \bottomrule
  \end{tabular}
  \end{table}

  \paragraph{Ball 3D parameters.}
  Per-trajectory parameters are wall restitution $e$, gravity $g$, and
  initial-velocity magnitude $\|v_0\|$; angular-velocity components are sampled
  uniformly in $[-5,\,5]\,\mathrm{rad/s}$ on every split. OOD shifts touch
  restitution and gravity only; the initial-velocity distribution is held fixed
  across all splits.

  \begin{table}[!htb]
  \centering
  \caption{Ball 3D per-split parameter sampling.
  $\|v_0\| \in [1,\,3]\,\mathrm{m/s}$ and
  $\boldsymbol{\omega}_0 \in [-5,\,5]^3\,\mathrm{rad/s}$ on all splits.}
  \label{tab:ball3d_params}
  \small
  \setlength{\tabcolsep}{4pt}
  \renewcommand{\arraystretch}{1.15}
  \begin{tabular}{l c c}
  \toprule
  Split & Restitution $e$ & Gravity $g$ (m/s$^2$) \\
  \midrule
  train / val / test & $[0.70,\,0.95]$ & $[-10.5,\,-9.0]$ \\
  OOD-near           & $[0.50,\,0.70]\,\cup\,[0.95,\,0.99]$ & $[-12.0,\,-10.5]\,\cup\,[-9.0,\,-7.5]$ \\
  OOD-far            & $[0.30,\,0.50]$ & $[-15.0,\,-12.0]\,\cup\,[-7.5,\,-5.0]$ \\
  \bottomrule
  \end{tabular}
  \end{table}

  \subsection{Training hyperparameters}
  \label{app:training_hp}

  All three surrogates are trained with AdamW, multi-horizon supervision over
  $h \in \{1, 2, 4, 8, 16, 32, 64\}$, and a $10\%$ DAgger refinement against
  the reference solver. Per-environment hyperparameters are summarised in
  Table~\ref{tab:training_hp}. We use one seed for the main-paper numbers and
  three seeds for the cross-seed analysis (Section~\ref{app:cross_seed}); no
  hyperparameters were tuned per seed. We use early stopping on validation
  MSE with patience $15$ epochs.

  \begin{table}[!htb]
  \centering
  \caption{Training hyperparameters per environment. ``Step'' counts a single
  gradient update; ``samples'' counts the number of $(s_0, s_T, T)$ triples
  seen during training.}
  \label{tab:training_hp}
  \small
  \setlength{\tabcolsep}{6pt}
  \renewcommand{\arraystretch}{1.15}
  \begin{tabular}{l c c c}
  \toprule
   & Oregonator & Euler 2D & Ball 3D \\
  \midrule
  Backbone & U-Net (4-stage) & U-Net (4-stage) & FiLM-MLP (4 blocks) \\
  Channel multipliers & $[1,\,2,\,4,\,4]$ & $[1,\,2,\,4,\,4]$ & n/a \\
  Base channels       & $48$ & $48$ & n/a \\
  Hidden dim          & n/a & n/a & $256$ \\
  Time embedding dim  & $64$ & $64$ & $64$ \\
  Parameters          & $3.56$\,M & $3.56$\,M & $0.69$\,M \\
  \midrule
  Optimiser           & AdamW & AdamW & AdamW \\
  Learning rate       & $2{\times}10^{-4}$ & $2{\times}10^{-4}$ & $3{\times}10^{-4}$ \\
  Weight decay        & $10^{-4}$ & $10^{-4}$ & $10^{-4}$ \\
  Gradient clip       & $1.0$ & $1.0$ & $1.0$ \\
  Warmup epochs       & $3$ & $0$ & $0$ \\
  LR schedule         & cosine & cosine & cosine \\
  Batch size          & $8$ & $8$ & $128$ \\
  Samples / epoch     & $5{,}000$ & $2{,}000$ & $25{,}600$ \\
  Epochs              & $60$ & $40$ & $80$ \\
  Total samples seen  & $300{,}000$ & $80{,}000$ & ${\approx}\,2{\times}10^{6}$ \\
  DAgger weight $\lambda$ & $0.1$ & $0.1$ & $0.1$ \\
  Early-stopping patience & $15$ & $15$ & $15$ \\
  Loss & MSE on normalised state & MSE on normalised state & MSE on normalised state \\
  \bottomrule
  \end{tabular}
  \end{table}

  \paragraph{Recipe choice.}
  The same hyperparameters are used across all three environments without
  per-environment tuning. We do not claim these values are globally optimal for
  any single environment; they are the smallest set that produced a usable
  surrogate on each of the three within our compute budget. The trust-signal
  results in Section~\ref{sec:5_trust} are properties of the step-doubling
  probe under multi-horizon training, not of these specific numerical values:
  a better-tuned surrogate would have lower absolute RMSE, but the
  rank-ordering of trajectories by $\hat{e}$ versus true error would be
  unaffected.

  \subsection{Hardware}
  \label{app:hardware}

  All training, evaluation, and benchmarking in this paper is performed on a
  single consumer laptop with the configuration in Table~\ref{tab:hardware}.
  The same machine is used for the CPU and GPU benchmarks reported in
  Figure~\ref{fig:mode1}; all reported wall-clock times reflect realistic
  single-machine deployment, not cluster-scale resources.

  \begin{table}[!htb]
  \centering
  \caption{Hardware configuration used for all training and benchmarking.}
  \label{tab:hardware}
  \small
  \setlength{\tabcolsep}{8pt}
  \renewcommand{\arraystretch}{1.15}
  \begin{tabular}{l l}
  \toprule
  GPU         & NVIDIA GeForce RTX 5050 Laptop GPU, $8$\,GB VRAM \\
  CPU         & AMD Ryzen 9 8940HX (16 cores, $24$ threads) \\
  System RAM  & $16$\,GB \\
  PyTorch     & $2.9.0$ + CUDA $12.8$, cuDNN $9.10$ \\
  OS          & Ubuntu $22.04$ \\
  \bottomrule
  \end{tabular}
  \end{table}

  \paragraph{Thread configuration for benchmarks.}
  For the same-hardware CPU comparison (Figure~\ref{fig:mode1}, panel a), both
  the surrogate and the reference solver are run with
  \texttt{torch.set\_num\_threads(8)} and \texttt{OMP\_NUM\_THREADS=8} to
  prevent CPU-thread saturation from dominating the wall time. GPU benchmarks
  (panel b) use the surrogate on the GPU at the indicated batch size and
  report wall time against the same unbatched CPU solver as the baseline; we
  do not fork the comparison across multiple GPUs.

  \paragraph{Reproducibility.}
  All training configs, solver implementations, dataset-generation scripts,
  and evaluation protocols will be released on acceptance. The reference
  solvers are standard textbook implementations (HLL finite-volume for Euler
  2D, Strang-split implicit-explicit for Oregonator, semi-implicit Euler for
  Ball 3D) without bespoke optimisations; speedup numbers in
  Figure~\ref{fig:mode1} are intended to characterise the recipe's behaviour
  under realistic single-machine deployment, not to upper-bound what an
  aggressively engineered solver could achieve.

 \section{Broader Impact}
  \label{app:broader_impact}

  The two methodological pieces of this paper, a label-free per-trajectory trust signal and a deployment policy that defers uncertain trajectories to the reference physics, are most valuable
  when wrong predictions are expensive and ground truth is bounded by the cost of asking the physical world. Robotics world models are exactly that regime, and the most consequential
  downstream uses of the recipe are there.

  \paragraph{Robotics deployment.}
  Three concrete uses follow. \textit{Selective sensor framing}: instead of running the physical robot at every step, an agent runs a learned world model continuously and queries physical
  sensing only on frames $\hat{e}$ flags, trading training-data cost against simulator bias on a controlled knob (the analogue of $q$). \textit{Trust-aware control}: a world model used inside
   RL or model-based control can expose $\hat{e}$ alongside its prediction so the policy trusts the simulator where coherent and falls back to behaviour cloning or planning under uncertainty
  where it is not. \textit{Sim-to-real allocation}: multi-horizon training localises where the simulator's internal coherence breaks first, often the same states (contact, deformation, novel
  objects) where the real world departs from simulation; a small real-world rollout budget can then be directed at exactly those regions. Lifted to a scene-graph-conditioned predictor,
  $\hat{e}$ would acquire per-object granularity, making selective rollback meaningful at the level a robot policy consumes (objects, contact events, grasps) rather than at the level of pixel
   grids.

  \paragraph{Scope and risks.}
  We have not demonstrated the recipe on robotics; verifying it on action-conditioned scene prediction, contact-rich manipulation, and raw sensor inputs are distinct research programs, and
  the benefits above are conditional on that verification. The familiar risk is that a trust signal right on average but silently wrong on a small subset of states could give policies false
  confidence in regimes where they should defer; Section~\ref{sec:limitations} reports one such failure in our benchmarks (Ball 3D under far-OOD restitution and gravity at $h{=}16$, where
  AUROC drops below chance). Deploying in safety-critical settings requires characterising and bounding these failure modes for the specific environment.
%%%%%%%%%%%%%%%%%%%%%%%%%%%%%%%%%%%%%%%%%%%%%%%%%%%%%%%%%%%%

\clearpage

\end{document}